\documentclass{article}

\PassOptionsToPackage{numbers, compress}{natbib}

\usepackage[preprint]{neurips_2022}

\usepackage[utf8]{inputenc} 
\usepackage[T1]{fontenc}    
\usepackage{hyperref}       
\usepackage{url}            
\usepackage{booktabs}       
\usepackage{amsfonts}       
\usepackage{nicefrac}       
\usepackage{microtype}      
\usepackage{xcolor}         
\usepackage{amsmath} 
\usepackage{amssymb}
\usepackage{graphicx}
\usepackage[ruled]{algorithm2e}
\usepackage{float}
\usepackage{comment}
\usepackage{wrapfig}
\usepackage{caption}
\title{RENs: Relevance Encoding Networks}

\author{%
  Krithika Iyer\\
  Scientific Computing and Imaging Institute,\\
  School of Computing, University of Utah, Salt Lake City, UT, USA\\
  \texttt{krithika.iyer@utah.edu}\\
  \And
  Riddhish Bhalodia\\
  Scientific Computing and Imaging Institute,\\
  School of Computing, University of Utah, Salt Lake City, UT, USA\\
  \texttt{riddhishb@gmail.com}\\
  \And
  Shireen Elhabian\\
  Scientific Computing and Imaging Institute,\\
  School of Computing, University of Utah, Salt Lake City, UT, USA\\
  \texttt{shireen@sci.utah.edu}
}

\begin{document}

\maketitle

\begin{abstract}

The manifold assumption for high-dimensional data assumes that the data is generated by varying a set of parameters obtained from a low-dimensional latent space. Deep generative models (DGMs) are widely used to learn data representations in an unsupervised way. DGMs parameterize the underlying low-dimensional manifold in the data space using bottleneck architectures such as variational autoencoders (VAEs). The bottleneck dimension for VAEs is treated as a hyperparameter that depends on the dataset and is fixed at design time after extensive tuning. As the intrinsic dimensionality of most real-world datasets is unknown, often, there is a mismatch between the intrinsic dimensionality and the latent dimensionality chosen as a hyperparameter. This mismatch can negatively contribute to the model performance for representation learning and sample generation tasks. This paper proposes relevance encoding networks (RENs): a novel probabilistic VAE-based framework that uses the automatic relevance determination (ARD) prior in the latent space to learn the data-specific bottleneck dimensionality. The relevance of each latent dimension is directly learned from the data along with the other model parameters using stochastic gradient descent and a reparameterization trick adapted to non-Gaussian priors. We leverage the concept of DeepSets to capture permutation invariant statistical properties in both data and latent spaces for relevance determination. The proposed framework is general and flexible and can be used for the state-of-the-art VAE models that leverage regularizers to impose specific characteristics in the latent space (e.g., disentanglement). With extensive experimentation on synthetic and public image datasets, we show that the proposed model learns the relevant latent bottleneck dimensionality without compromising the representation and generation quality of the samples. 

\end{abstract}

\section{Introduction}
\label{introduction}

Due to the rapidly evolving computational technologies, large amounts of unlabeled data are continuously generated. Considerable time, labor, and resources are dedicated to labeling, pre-processing, and transforming the unlabelled data for real-world supervised machine learning applications. As an alternative, unsupervised representation learning algorithms extract meaningful and discriminative representations that are amenable to the downstream task in the absence of labels. Unsupervised representation learning have found applications in several domains, such as computer vision \cite{kim2019diversify,wang2020deep,lin2017marta,jahanian2021generative,kim2021hybrid}, medical image analysis \cite{tang2017medical,yadav2021lung,kolyvakis2018biomedical}, and  natural language processing \cite{han2021unsupervised,radford2018improving}

Deep generative models are widely used in unsupervised representation learning to learn informative representations of data and parameterize the underlying data manifold, enabling them to generate new samples from the distribution. Widely used methods include density estimation using flow-based models \cite{dinh2016density,zang2020moflow,stypulkowski2019conditional}, generative adversarial networks (GANs) \cite{tanaka2019data,goodfellow2014generative}, and variational autoencoders \cite{kingma2013auto, rezende2014stochastic}. This paper focuses on autoencoder-type architectures like VAEs that provide representation and sample generation capabilities. 

Algorithms and models for unsupervised representation are based on the manifold assumption, where the data is assumed to be generated by varying a set of parameters. Usually, the number of parameters is much smaller than the dimensionality of the data. For example, different facial images can be generated using a finite set of parameters such as lighting, skin color, expressions, facial features, hair, etc. According to the manifold assumption: a set of $N-$samples $\set{X} = \{\sample{x}_1, ..., \sample{x}_N\}$ in a $D-$dimensional data space, where \(\sample{x} \in \realdim{R}^D\), is said to lie on or near a low-dimensional manifold \(\mathcal{M} \subset \mathbb{R}^D\) with intrinsic dimensionality \(d_{\mathcal{I}}\ll D\) that refers to the minimum number of parameters necessary to capture the entire information content present
in the representation \cite{gong2019intrinsic} 

Principal component analysis (PCA) was one of the earliest methods that established a relation between the data and the low-dimensional latent space. PCA provides a closed-form solution for determining the optimum latent space dimensionality. However, PCA is limited by linearity and imposes oversimplifying assumptions of Gaussianity on the data distribution. Moreover, traditional approaches such as PCA are not scalable for large datasets and require algorithmic treatments (e.g., online PCA \cite{cardot2018online,boutsidis2014online}) for big data. Deep generative models overcome the limitations of PCA providing non-linearity and scalability using deep networks to parameterize the mapping to the latent space. In the case of deep generative models, the dimensionality of the latent space is typically defined upfront for each dataset at the design time. The design process
may be under- or over-provision the number of dimensions for the application at hand. If the dimensionality is not predefined, this parameter is usually determined using time and resource-consuming cross-validation. A mismatch between latent dimensionality $L$ and intrinsic dimensionality \(d_\mathcal{I}\) affects the quality of data representation and sample generation \cite{rubenstein2018latent,mondal2021flexae,rubenstein2018wasserstein}. Studying and understanding the effects of dimensionality mismatch in the latent space becomes imperative to use deep generative models effectively.

\begin{wrapfigure}[18]{R}{0.5\textwidth}
    \centering
     \vspace{-0.1in}
    \includegraphics[width=0.5\textwidth]{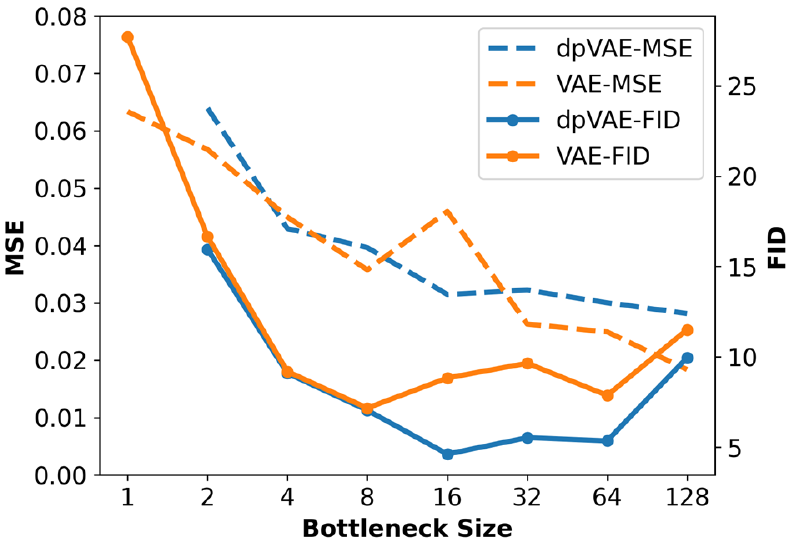} \vspace{-0.2in}
    \caption{\textbf{Larger latent dimensions are not always better.} Frechet inception distance (FID) scores (\textit{lower is better}) and reconstruction mean square error (MSE, \textit{lower is better}) of VAE and dpVAE \cite{bhalodia2020dpvaes} models with varying latent dimensionality for MNIST. }
    \label{fig:teaser_exp}
\end{wrapfigure}

Multiple studies have targeted improving reconstruction quality, but very few have tried to analyze the impact of latent dimensionality mismatch. Wasserstein autoencoders, highlights that a mismatch between the latent dimensionality and the true intrinsic dimensionality leads to an infeasible optimization object \cite{mondal2021flexae,mondal2019maskaae}. For deterministic encoders and decoders, a high capacity bottleneck can cause curling of the manifold \cite{tolstikhin2017wasserstein}. Whereas using a smaller bottleneck can cause lossy compression of data and deteriorate representation quality \cite{de2020dynamic}. For deterministic encoders, studies have concluded that larger bottleneck dimensions are not always better \cite{mondal2019maskaae,tolstikhin2017wasserstein}. For VAE, it was theoretically proved that increasing the bottleneck capacity beyond the intrinsic dimensionality does not improve the reconstruction quality \cite{dai2019diagnosing}. 

We propose a principled framework grounded in probabilistic modeling to identify the optimal data-specific latent dimensionality without adding new hyperparameters. To empirically motivate the proposed model, we performed experiments with - vanilla VAE \cite{kingma2019introduction} and dpVAE \cite{bhalodia2020dpvaes} to analyze the impact of latent dimensionality on the representation learning and sample generation tasks. See Figure~\ref{fig:teaser_exp}. The two models were trained on the MNIST dataset with varying latent dimensions, while other parameters were kept the same. The Frechet inception distance (FID) score was used as the generation metric, and mean squared error (MSE) as the representation metric. The FID score and MSE curve indicate inferior performance when the bottleneck size is under-provisioned. We see the MSE performance improve with the bottleneck size, but FID scores suffers at higher dimensions such as 64 and 128. The lower MSE and increase in FID score at a large bottleneck size indicates over-fitting and loss of generalization. We can conclude that larger dimensions are not always guaranteed to produce the best-performing models. Hence, we need automated ways of identifying the dimensionality mismatch and informing the model about the intrinsic dimensionality of a given dataset. 
Our contributions are as follows:
\begin{enumerate}
    \item Introduce relevance encoding networks (RENs): a framework that facilities the training of VAEs using a unified formulation to parameterize the data distribution and detect latent-intrinsic dimensionalities mismatch. The formulation is general and can be adapted to state-of-art VAE-based methods such as regularized VAEs. This framework also provides a PCA-like ordering for the latent dimensions that conveys the variance of each latent dimension supported by the data manifold.
    \item Derive the evidence lower bound (ELBO) for RENs in case of vanilla VAEs and decoupled prior VAEs \cite{bhalodia2020dpvaes} that leverages invertible bottleneck to improve the matching of the aggregate posterior with the latent prior.
    \item Use \(\sigma\)-VAE \cite{rybkin2021simple} to calibrate the RENs decoder and reduce the requirement of tuning the weight on the likelihood term of the VAE ELBO.
    
    \item Demonstrate the ability of relevance encoding networks in detecting the relevant bottleneck dimensionality for three public image datasets without compromising the representation and generation quality and with no additional hyperparameter tuning.
\end{enumerate}

\section{Related Work}

VAE \cite{kingma2019introduction} is a latent variable model specified by an encoder, decoder, and prior distribution on the latent space. The encoder maps the input to the latent space (inference), while the decoder reconstructs the original input from the latent space (representation). The prior enables sample generation from a tractable probabilistic distribution \cite{doersch2016tutorial}. Several studies have suggested that using learnable prior can improve the performance of VAEs and reduce the impact of dimensionality mismatch \cite{xu2020shallow,xu2019necessity,tomczak2018vae,bhalodia2020dpvaes,xu2019necessity}.
Dai and Wipf rigorously analyzed the VAE objective under various scenarios of dimensionality mismatch \cite{dai2019diagnosing}. A critical conclusion (see Theorem 5 in \cite{dai2019diagnosing}) states that optimal reconstruction can be achieved when the latent bottleneck dimensionality matches the intrinsic dimensionality, and increasing the bottleneck capacity may negatively impact the generation process.

Very few methods have demonstrated ways of identifying and handling latent-intrinsic dimensionality mismatch. De Boom et al. illustrated the use of Generalized ELBO with constrained optimization (GECO) and the L0-augment-REINFORCE-merge (L0-ARM) gradient estimator \cite{li2019l_0,yin2018arm} to shrink the latent bottleneck dimensionality of VAEs automatically \cite{de2020dynamic}. The L0 norm was applied to a global binary gating vector that controlled the latent dimensionality. GECO was used to define a constraint on the reconstruction error to give it more weight during optimization until the desired level of accuracy is reached. Once the threshold is reached, the narrowing of the bottleneck is given priority. Kim et al. proposed relevance factor VAE \cite{kim2019relevance} that infers relevance and disentanglement using total correlation. Although the models proposed by De Boom et al. and Kim et al. identify the inactive latent dimensions and eliminate them in the variational posterior distribution, the prior is still an isotropic standard normal. This can result in poor representation and generation quality.

Heim N et.al. proposed the relevance determination in differential equations model (Rodent) model \cite{heim2019rodent} that showed the use of automatic relevance determination (ARD) \cite{tipping1999relevance,bishop2013variational} priors to minimize the state size of the ordinary differential equation (ODE) and the number of non-parameters required to solve the problem using partial observations. They used a VAE-like architecture, where the encoder was a neural network, and the decoder was an ODE solver. Isotropic Gaussian was used with an ARD prior in the latent space, and a point estimate was used for the variance of the prior distribution. The Rodent model formulation is not fully probabilistic and only focused on solving the ODEs.

For autoencoders, several studies (e.g., \cite{rubenstein2018latent,rubenstein2018wasserstein}) have analyzed how deterministic and random encoder-decoder pairs perform in the presence of latent-intrinsic dimensionality mismatch. Studies by Rubenstein et al. revealed that the deterministic encoders start curling the manifold in the latent space when the latent dimensionality is higher than the intrinsic dimensionality. At the same time, random encoders fill the irrelevant dimension with noise while encoding useful information in the latent space. Random encoders start behaving like deterministic encoders if the dimensionality is increased further. Deterministic and random exhibit poor sample generation performance with an increase in the volume of the holes in the latent space. Mondal et al. studied the effect of dimensionality mismatch in the case of deterministic autoencoders \cite{mondal2019maskaae,mondal2021flexae}. Mathematically and empirically, Mondal et al. shows that having a fixed prior distribution, oblivious to the dimensionality of the true latent space, leads to the optimization infeasibility, and proposes masked autoencoders (MAAE) \cite{mondal2019maskaae} as a potential solution. MAAE \cite{mondal2019maskaae} introduced modifications to the autoencoder architecture to infer a mask at the end of the encoder to suppress noisy latent dimensions.

Existing approaches that identifies the relevant dimensions, introduces more hyperparameters that have to be tuned to identify the bottleneck size. Hence, the complexity of finding the optimum latent dimension remains the same. Moreover, these methods treat relevance determination as a separate task agnostic to the probabilistic formulation of deep generative models, making the solution less interpretable.The RENs framework facilities training of VAEs using a unified probabilistic formulation to parameterize the data distribution and detect latent-intrinsic dimensionality mismatch without adding new hyperparameters.

\section{Background}\label{background}

\textbf{Notation:} We denote a set of \(N-\)observations in a $D-$dimensional data space as \(\set{X} = \{\sample{x}_1, ...,\sample{x}_N\}\) and their corresponding latent representations \(\set{Z} = \{\sample{z}_1, ...,\sample{z}_N\}\). A representation learning model maps an observation \( \sample{x}_n \in \realdim{R}^D\) to an unobserved latent representation \(\sample{z}_n \in \realdim{R}^L\) in an $L-$dimensional latent space, where $L \ll D$. Hereafter, we use boldface lowercase letters to denote vectors, boldface uppercase letters to denote matrices, and non-bold lowercase to denote scalars. 

\textbf{Variational Autoencoders (VAEs):} VAEs are latent variable models that learn data representations in an unsupervised way by matching the learned model distribution \(p_{\theta}(\sample{x})\) to the true data distribution \(p(\sample{x})\). The generative (i.e., decoder) and inference (i.e., encoder) models in VAEs are jointly trained to maximize a tractable lower bound \(\mathcal{L}(\theta,\phi)\) on the marginal log-likelihood \(\mathbb{E}_{p(\sample{x})} [\log p_{\theta}(\sample{x})]\) of the training data. The structure of the learned latent representation is controlled via imposing a prior distribution on the latent space such as \(p(\sample{z}) \sim \mathcal{N}(\sample{z}; \mathbf{0},\mathbb{I})\).
\begin{equation}\label{vae_eq}
    \mathcal{L}(\theta,\phi) = \mathbb{E}_{p(\sample{x})} \left[\mathbb{E}_{q_{\phi}(\sample{z} | \sample{x})} [\log p_{\theta}(\sample{x}|\sample{z})]  - \operatorname{KL} [q_{\phi}(\sample{z} | \sample{x}) || p(\sample{z})]\right]
\end{equation}
where \(\theta\) denotes the generative model parameters, \(\phi\) denotes the inference model parameters, and \(q_{\phi}(\sample{z}|(\sample{x}) \sim \mathcal{N}\left(\sample{z}; \boldsymbol{\mu}_{\sample{z}}(\sample{x}),\boldsymbol{\Sigma}_{\sample{z}}(\sample{x})\right)\) 
is the variational posterior distribution that approximates the true posterior 
\(p(\sample{z}|\sample{x})\), with 
\( \boldsymbol{\mu}_{ \sample{z}}(\sample{x}) \in \realdim{R}^{L}, \boldsymbol{\Sigma}_{ \sample{z}}(\sample{x}) = 
\operatorname{diag}(\boldsymbol{\sigma}_{\sample{z}}(\sample{x}))\), and \(\boldsymbol{\sigma}_{\sample{z}}(\sample{x}) \in \mathbb{R}^{L}_+\)

\textbf{dpVAE: Decoupled Prior for VAE:} Maximizing the ELBO by optimizing the marginal log-likelihood does not guarantee good representation. With expressive generative models \(p_{\theta}(\sample{x}|\sample{z})\), VAE can ignore the latent representation and not encode any information about the data in them, but still maximize the ELBO \cite{hoffman2016elbo,alemi2018fixing,chen2016variational}, this is the information preference phenomena of VAEs. It has been shown that data-driven (i.e., learned during training) priors help mitigate the information preference of VAEs \cite{hoffman2016elbo,rosca2018distribution,xu2019necessity}. Specifically, dpVAE \cite{bhalodia2020dpvaes} decouples the latent space that performs the representation \(\sample{z}\) and the space that drives sample generation \(\sample{z}_0\) using a functional bijective mapping \(g_{\eta}(\sample{z}) = \sample{z}_0\) parameterized by the network parameters \(\eta\), where \(p(\sample{z}_0) \sim \mathcal{N}(\sample{z}_0;\mathbf{0},\mathbb{I})\). Affine coupling layers \cite{dinh2016density} are used to build a flexible bijection function \(g\) by stacking a sequence of \(K\) simple bijection blocks. 
\begin{equation}\label{dpvae_eq}
    \begin{split}
     \mathcal{L}_{\operatorname{dpVAE}}(\theta,\phi) &=  \mathbb{E}_{p(\sample{x})} [\mathbb{E}_{q_{\phi}(\sample{z} | \sample{x})} [\log            p_{\theta}(\sample{x}|\sample{z})] 
       + \mathbb{E}_{q(\sample{z}|\sample{x}} \left[ \sum^{K}_{k=1} \sum^{L}_{l=1} b^{l}_k s_k (b^{l}_k z^{l}_k) \right]\\
     & -\frac{1}{2} \log|\boldsymbol{\Sigma}_{ \sample{z}}(\sample{x})| 
       + \mathbb{E}_{q(\sample{z}|\sample{x}}\left[g_{\eta}(\sample{z})^T g_{\eta}(\sample{z}\right)]
    \end{split}
\end{equation}
where \(\eta = \{s_1,...,s_k, t_1,...t_k\}\), \(s_k\) is the scaling network of the \(k-\)th block, \(b^{l}_k\) is the \(l-\)th element in \(\sample{b_k}\) vector; the binary mask used for partition the \(k-\)th block of the scaling \(s_k\) and translation function \(t_k\) of the \(k-\)th block and \( z^{l}_k\) is the \(k-\)th element of \(\sample{z_k}\) vectors

\textbf{Sigma Variational Autoencoders:} It is common practice to consider the decoding distribution \(p_{\theta}(\sample{x}|\sample{z})\) as a Gaussian with constant variance representing the data noise (tuned as a hyperparameter). When using a fixed variance, a model with high variance will not retain enough information in the latent space to faithfully reconstruct samples and a model with low variance will generate poor samples as the \(\operatorname{KL}\) divergence term becomes weaker \cite{alemi2018fixing,lucas2019don}. The \(\sigma\)-VAE \cite{rybkin2021simple} model is a simple, yet effective solution for calibrating the decode variance by using  a single learnable parameter \(\sigma\) in \(p_{\theta,\sigma}(\sample{x}|\sample{z}) = \mathcal{N}(\sample{x}; \mu_{\sample{z}}(\sample{x}),\sigma^2 \mathbb{I}) \)
The variance of the decoder is trainable and is learned along with the rest of the model parameters \cite{rybkin2021simple}. The formulation reduces the time and resources required to tune the variance for each model.

Considering the performance merits of dpVAE and \(\sigma\)-VAE, all our experiments will implement the VAE model with the decoupling architecture of dpVAE and \(\sigma\)-VAE formulation. This dovetails with RENs objective to reduce VAE hyperparamters that require extensive tuning for each datasets while improving sample generation and representation.

\section{Relevance Encoding Networks}

  Larger latent bottleneck sizes does not guarantee a better VAE performance (Figure~\ref{fig:teaser_exp}). Therefore, to inform the model about intrinsic dimensionality, we introduce an automatic relevance determination (ARD) \cite{bishop2013variational,tipping1999relevance} hyperprior over the latent space prior \(p(\mathbf{z}_n)\). 
 
\begin{wrapfigure}[8]{R}{0.57\textwidth}
    \centering
    \vspace{-0.5in}
    \includegraphics[scale=0.55]{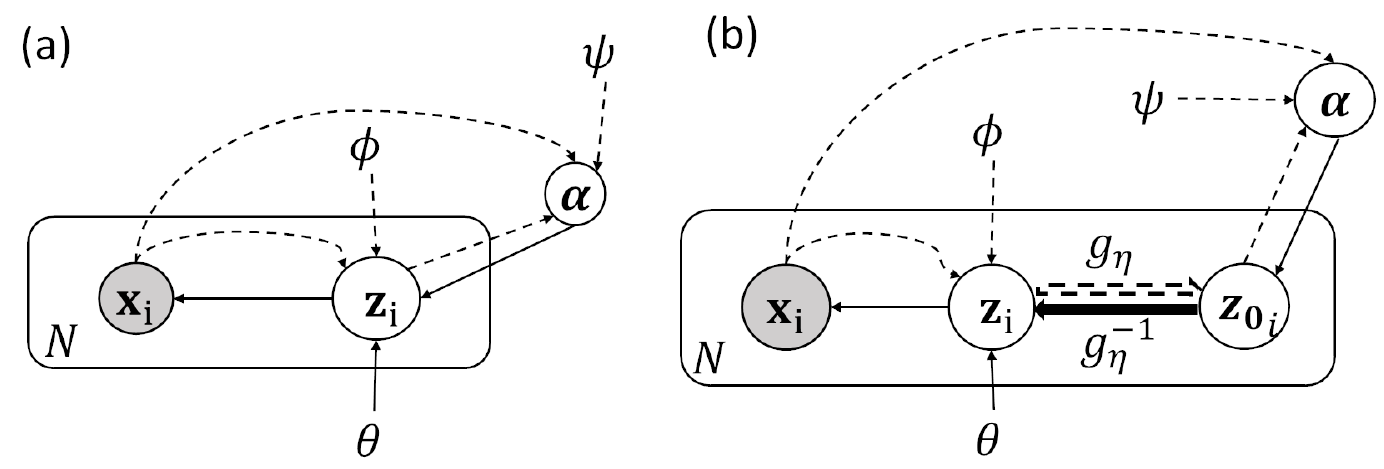}
    \caption{ Graphical models for (a) REN with VAE (b) REN with dpVAE. The solid lines indicate the variational inference flow and the dotted lines indicate the generative flow for all the models. The block arrows indicate the invertible flow network.}
    \label{fig:graphical_model}
\end{wrapfigure}
 
\subsection{RENs Formulation}\label{ren_formulation}

The ARD hyperprior regularizes the latent space to discover relevant latent dimensions that are supported by the data, hence reducing the contribution of redundant dimensions. The ARD hyperprior provides relevance of each dimension in the latent representation and this relevance is defined via precision (i.e. inverse of variance). The ARD hyperprior pushes the precision of the spurious dimensions to infinity; thus, the variance of these dimensions is pushed to zero in the latent space. 
The latent prior is given by:
\begin{equation}\label{ard_prior_1}
    p(\set{Z}|\boldsymbol{\alpha}) = \prod_{n=1}^N p(\sample{z}_n|\boldsymbol{\alpha}) \quad \textrm{where} \quad p(\sample{z}_n|\boldsymbol{\alpha}) \sim \mathcal{N}(\sample{z}_n; \mathbf{0},\boldsymbol{\alpha}^{-1}\mathbb{I}) 
\end{equation}
with ARD hyperprior given as:
\begin{equation}\label{ard_prior_2}
    p(\boldsymbol{\alpha}) = \Gamma(a \mathbf{1}_L,b\mathbf{1}_L)
\end{equation}
Here, \(\boldsymbol{\alpha} \in \realdim{R}_+^L\) is the relevance of the latent dimensions, and \(\mathbf{1}^L \) is an $L-$dimensional vector of ones. The concentration parameter \(a  \in \mathbb{R}_+\) and the rate parameter \(b \in \mathbb{R}_+\) of the Gamma distribution are shared across all latent dimensions.
The VAE prior now becomes:
\begin{equation}
    p(\set{Z}) = \left[\prod_{n=1}^N p(\mathbf{z}_n|\boldsymbol{\alpha})\right]p(\boldsymbol{\alpha}) = \left[\prod_{n=1}^N \mathcal{N}(\sample{z}_n;\mathbf{0},\boldsymbol{\alpha}^{-1}\mathbb{I})\right]\Gamma(a \mathbf{1}_L,b\mathbf{1}_L) 
\end{equation}

\begin{figure}
    \centering
    \includegraphics[scale=0.9]{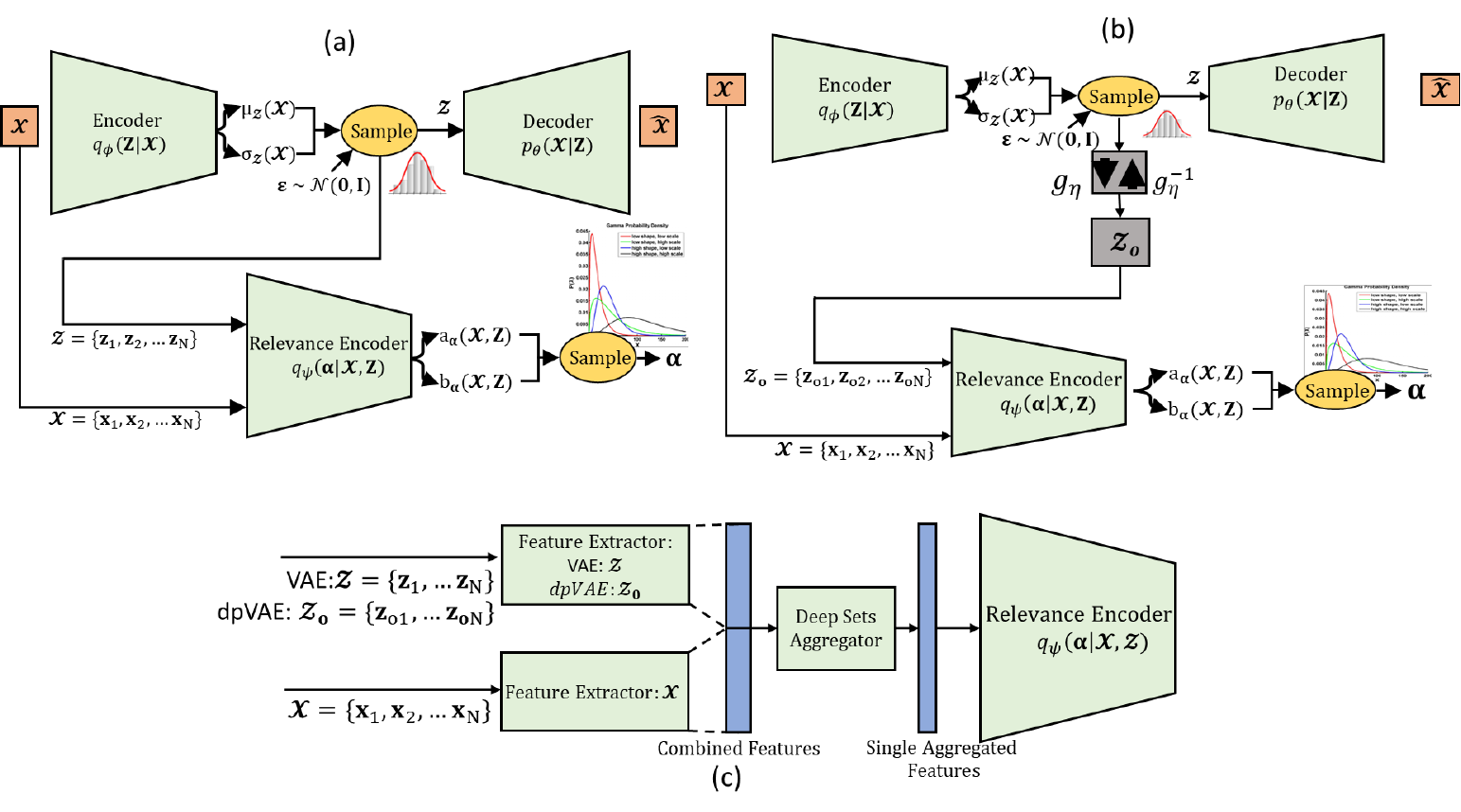}
    \caption{Block Diagram of (a) Relevance encoding networks (REN) with VAE, (b) REN with dpVAE. The model consists of encoder(\(\phi\)), decoder(\(\theta\)), and relevance encoder(\(\psi\)). The relevance encoding network infers the variational posterior \(q_{\psi}(\boldsymbol{\alpha}|\set{X},\set{Z})\) (c) The relevance encoder is broken down into its constituent parts. It consists of two feature extracts: one for the data and the other for the latent representation. The features are combined and passed to the deep sets aggregator. The aggregated feature is fed to the final relevance encoder that approximates the concentration and rate of the gamma distribution.}
    \label{fig:ren_block_diagram}
\end{figure}

We introduce a relevance encoder to the VAE architecture that learns the variational posterior \(q_{\psi}(\boldsymbol{\alpha}|\set{X},\set{Z}) = \Gamma(a_{\boldsymbol{\alpha}}(\set{X},\set{Z}), b_{\boldsymbol{\alpha}}(\set{X},\set{Z}))\) which approximates the true posterior \(p(\boldsymbol{\alpha}|\set{X},\set{Z})\). Here, $\psi$ denotes the parameters of the relevance encoder network. The relevance of a latent dimension is a statistical property of the underlying latent distribution that is induced by the data distribution. Consequently, relevance cannot be estimated from a single sample, but instead requires access to a finite set of representative samples for the data and latent distributions. Hence, we formulate a set-input problem, where a set of instances is given as an input and the relevance encoder parameterizes the relevance for the entire set with permutation invariance. Taking the complete dataset into consideration, the joint probability of the training data can be expressed as, \( p(\set{X},\set{Z},\boldsymbol{\alpha}) = p(\set{X}|\set{Z},\boldsymbol{\alpha}) p(\set{Z}|\boldsymbol{\alpha})p(\boldsymbol{\alpha})\) and the probability of the training data is \(p(\set{X}) = \prod_{n=1}^N p(\sample{x}_n)\). 
See Figure~\ref{fig:graphical_model} and  Figure~\ref{fig:ren_block_diagram} for the plate notation of the graphical model and the block diagrams of their architectures, respectively. 

Samples of the Gamma distribution are reparameterized to enable gradient flow and network training in the presence of probabilistic layers. The derivatives are computed using the implicit reparameterization approach \cite{figurnov2018implicit}. This reparameterization is implemented in the TensorFlow Probability\footnote{https://www.tensorflow.org/probability/api\_docs/python/tfp/distributions/Gamma}.

Derivations for the ELBO can be found in Appendix~\ref{elbo_proof}. The ELBO for VAE is:
\begin{equation}\label{vae_ren_eq}
    \begin{split}
        \mathcal{L}(\theta, \phi, \psi) &= \mathbb{E}_{p(\set{X})} [\mathbb{E}_{q(\set{Z},\boldsymbol{\alpha} | \set{X})} [\log p_{\theta}(\set{X}|\set{Z},\boldsymbol{\alpha})]  \\
        &-\mathbb{E}_{q(\set{Z},\boldsymbol{\alpha} | \set{X})}[\log q_{\phi}(\set{Z}|\set{X}) ] 
        - \mathbb{E}_{q(\set{Z},\boldsymbol{\alpha} | \set{X})}[\log q_{\psi}(\boldsymbol{\alpha}|\set{X}.\set{Z})] \\ 
        &+\mathbb{E}_{q(\set{Z},\boldsymbol{\alpha} | \set{X})}[\log p(\set{Z}|\boldsymbol{\alpha})] 
        + \mathbb{E}_{q(\set{Z},\boldsymbol{\alpha} | \set{X})}[\log p(\boldsymbol{\alpha})] 
    \end{split}
\end{equation}
The ELBO for dpVAE is:
\begin{equation}\label{dpvae_ren_eq}
    \begin{split}
        \mathcal{L}(\theta, \phi, \psi) &= \mathbb{E}_{p(\set{X})} [\mathbb{E}_{q(\set{Z},\boldsymbol{\alpha} | \set{X})} [\log p_{\theta}(\set{X}|\set{Z},\boldsymbol{\alpha})]  \\
        &-\mathbb{E}_{q(\set{Z},\boldsymbol{\alpha} | \set{X})}[\log q_{\phi}(\set{Z}|\set{X}) ] 
        - \mathbb{E}_{q(\set{Z},\boldsymbol{\alpha} | \set{X})}[\log q_{\psi}(\boldsymbol{\alpha}|\set{X}.\set{Z})] \\ 
        & + \mathbb{E}_{q(\set{Z},\boldsymbol{\alpha} | \set{X})}[\log p(\boldsymbol{\alpha})]
        + \mathbb{E}_{q(\set{Z},\boldsymbol{\alpha} | \set{X})} [\log p(g_{\eta}(\set{Z})) 
         + \log (\begin{vmatrix}\frac{\partial g_{\eta}(\set{Z})}{\partial \set{Z}}\end{vmatrix})] 
      \end{split}
\end{equation}       

\subsection{Network and Training Strategies}
The ideal scenario would be feeding the relevance encoders all the training samples at once to generate a robust estimate of the relevance. However, the use of the entire dataset to estimate such relevance negatively impact the scalability provided by stochastic gradient descent training for large datasets. We train RENs via stochastic gradient descent and alternating optimization with two batch sizes. Each batch in the training dataset is broken down into \(r\) smaller batches of size \(batch\_size/r\). The alternating optimization is given by following steps: 
(i) The relevance encoder is kept fixed (i.e., not-trainable) and the smaller batches are used to update the VAE encoder \(\phi\)  and decoder \(\theta\) \(r\)-times by keeping the \(\boldsymbol{\alpha}\) obtained from the previous iteration fixed. (ii) The original large batch is used to update the entire network end-to-end (i.e., encoder \(\phi\), decoder \(\theta\), and relevance encoder \(\psi\)), and the \(\boldsymbol{\alpha}\) value is updated. As mentioned in section~\ref{ren_formulation}, we use set formulation for the relevance encoder \(\psi\), wherein the relevance encoder is fed in a set of data samples, and their latent representations and a single \(\boldsymbol{\alpha}\) is inferred. The response of the relevance encoder should be invariant to the ordering of the samples in the given batch. Therefore, we use DeepSets \cite{zaheer2017deep} to make the relevance encoder permutation invariant for a given batch. (see Figure~\ref{fig:ren_block_diagram}(c)). 
Using this alternating optimization and DeepSets aggregator, REN is encouraged to learn the global statistics of latent representations induced by the data distribution in the data space. See Algorithm.\ref{ren_algo} for more details.


\section{Experiments}\label{experiments}

\subsection{Toy Datasets}\label{toy_exp}
We use the circle and one-moon datasets for proof-of-concept experiments, both of which exhibit an intrinsic dimensionality of one, parameterized by the radius. We generated three different datasets for circle and one-moon distributions by varying the standard deviation of the zero-mean additive Gaussian noise to mimic data with different noise levels. For all our experiments, we implemented the relevance encoder with dpVAE \cite{bhalodia2020dpvaes} and \(\sigma\)-VAE framework \cite{dai2019diagnosing}, hereafter referenced as R-dpVAE. We tested the ability of the models to identify the intrinsic dimensionality along with their sample reconstruction capacity and realistic sample generation capability. The R-dpVAE model was compared with relevance factor VAE (RF-VAE) \cite{kim2019relevance} and masked adversarial autoencoders (MAAE) \cite{mondal2019maskaae}.  

\begin{figure}[!t]
    \centering
    \includegraphics[scale=0.88]{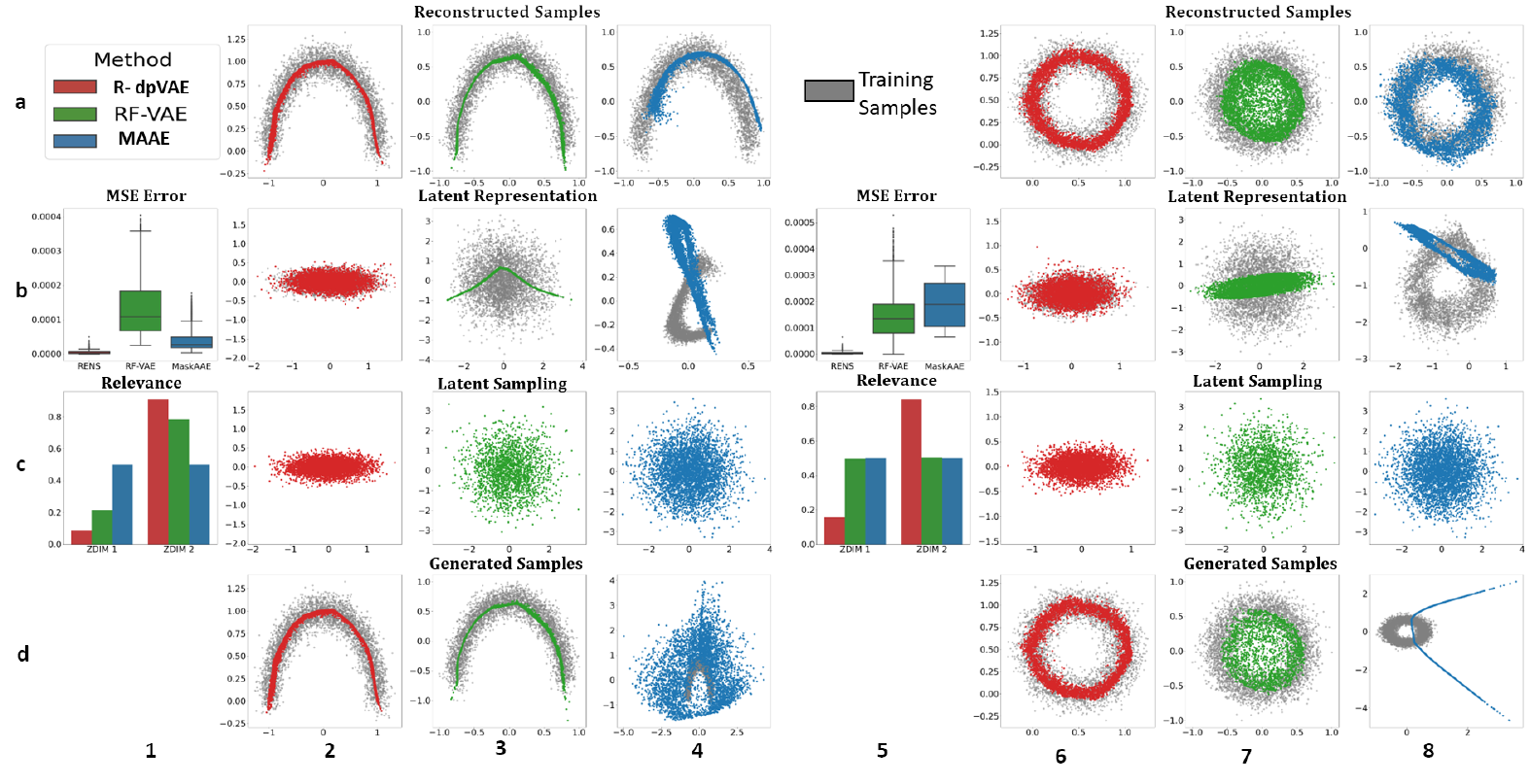}
    \caption{Reconstruction and sample generation for one-moon and circle at 10\% noise level.}
    \label{fig:one_moon_circle_0.01}
\end{figure}

Figure~\ref{fig:one_moon_circle_0.01} shows the results for one-moon and circle datasets with additive Gaussian noise of zero-mean and 10\% of the radius of the data manifold as standard deviation. Compared to RF-VAE and MAAE, R-dpVAE can regularize the latent space to discover latent dimensions relevance supported by the data while achieving the lowest mean square errors on the testing samples and suppressing the spurious dimension (Figure~\ref{fig:one_moon_circle_0.01}.2 and \ref{fig:one_moon_circle_0.01}.6). For R-dpVAE, the variance in the latent space is indicative of the relevance shown in plots (Figure~\ref{fig:one_moon_circle_0.01}.1c and \ref{fig:one_moon_circle_0.01}.5c). For one-moon, relevance estimated by R-dpVAE is \([0.2, 0.8]\) for zdim1 and zdim2, which correspond to the x and y-axis in the latent space. Therefore, the x-axis with low relevance has a larger variance, and the y-axis with high relevance has a low variance, correctly capturing a low-dimensional manifold where latent dimensionality equals the intrinsic dimensionality.

Although, the performance of RF-VAE (Figure~\ref{fig:one_moon_circle_0.01}.3 and \ref{fig:one_moon_circle_0.01}.7) comes close to R-dpVAE with the estimated relevance, it fails to generate good quality samples as the relevance is not factored in the aggregate posterior, and the latent prior is still a standard normal distribution. MAAE (Figure~\ref{fig:one_moon_circle_0.01}.4 and \ref{fig:one_moon_circle_0.01}.8) identifies the latent manifold but has a higher reconstruction error and generates bad quality samples due to the weak nature of regularization in the latent space; holes can be seen in the latent space (Figure~\ref{fig:one_moon_circle_0.01}.4b and \ref{fig:one_moon_circle_0.01}.8b). Across all experiments, R-dpVAE models provide a tighter distribution for the reconstruction error than other methods. R-dpVAE is the best performing model consistently, even in the presence of higher noise levels that make learning the underlying one-dimensional manifold more challenging. Results with different noise levels can be found in Appendix~\ref{appendix_toy_results}.\\

\vspace{-1pt}
\subsection{Image Dataset}\label{image_experiments}

We experimented with three image datasets: MNIST, Fashion MNIST, and dSprites. Similar to the toy experiments, we compared the performance of R-dpVAE with relevance factor VAE (RF-VAE) \cite{kim2019relevance}, masked adversarial autoencoders (MAAE) \cite{mondal2019maskaae}, with the addition of VAE regularized with L0 ARM and GECO (henceforth referenced as GECO) \cite{de2020dynamic}. To set baselines for comparisons, we implemented vanilla VAE with \(\sigma\)-VAE framework and dpVAE with \(\sigma\)-VAE framework sans the relevance. The supplementary material includes details on all the models' implementation, architectures, and hyperparameters. We consider the following quantitative metrics to evaluate and compare the models:  (1) \textbf{Frechet Inception Distance (FID)}: This metric calculates the distance between feature vectors of the real and generated images \cite{heusel2017gans}. Lower FID scores are better. (2) \textbf{Mean Squared Error (MSE):} The MSE values reported for reconstructed images are averaged over dimensions and sample size. (3) \textbf{Latent Dimensionality ($L^*$): } The estimated latent bottleneck size $L^*$ and  $\max(L^*) = L$, where $L$ is the provisioned bottleneck size of the model.

Although the ground truth intrinsic dimensionality of image datasets is not known, we use previous studies \cite{kim2019relevance,mondal2019maskaae,mondal2021flexae,de2020dynamic} as points of reference. The design choices of latent dimensions in the experiments are also motivated by the reported latent dimensions in the relevant literature. For MNIST, studies reported the latent dimensionality between 7 and 10. Hence, we choose 16 as the base dimensionality to provide the models with enough degrees-of-freedom to discover the relevant latent dimensions and 32 as the over-provisioned model to assess the impact of a significant mismatch with the intrinsic dimensionality. Similarly, for Fashion MNIST, the choices were 32 and 64 for base and over-provisioned dimensionalities, and for dSprites (known to have 6 factors of variation \cite{dsprites17}), the choices were 10 and 15.
The MAAE and GECO models provide the number of active dimensions, whereas the RF-VAE model provides a relevance vector with values from 0 to 1. For RF-VAE, we estimate the dimensionality by calculating the number of dimensions with relevance values higher than the average of the vector. In the case of R-dpVAE, the relevance estimated by the relevance encoding network is the inverse variance as per Eq.~\ref{ard_prior_1}. We compute the explained variance as the ratio of variance in a single dimension to the sum of all the variances. The number of dimensions required to explain 95\% of the variability is considered the detected latent bottleneck dimensionality.

Table~\ref{tab_fid_gen} summarizes the FID scores of randomly generated images and the estimated latent dimensionality by each model for all three datasets. The proposed model (R-dpVAE) achieves lower FID scores on all three datasets compared to the other models while estimating bottleneck dimensionality in the same range as the other methods. R-dpVAE also achieves better FID scores than the baseline models that do not perform relevance determination. This performance boost via relevance encoding bolsters our argument for the necessity of identifying and fixing the latent and intrinsic dimensionality mismatch. Similar to the findings from the toy experiments, the models that do not modify the posterior based on the learned relevance of the latent space exhibit inferior sample generation. This is reflected in the higher FID scores of RF-VAE, GECO, and MAAE. The MAAE and GECO models show consistency in determining the effective bottleneck size in the case of MNIST and dSprites, irrespective of the provisioned bottleneck size of the model. Although the R-dpVAE model estimates the bottleneck size in the same range as MAAE and GECO and shows better generation capabilities, R-dpVAE does not estimate the same size across \(L\) for the same dataset. This behavior could be attributed to the use of random samples in the mini-batch used for updating the relevance encoder(\(\psi\)) and may not capture the entire data variation. Using a stratified sampling approach to generate mini-batch could be a potential solution.

\begin{figure}[ht!]
    \centering
    \includegraphics[width=\textwidth]{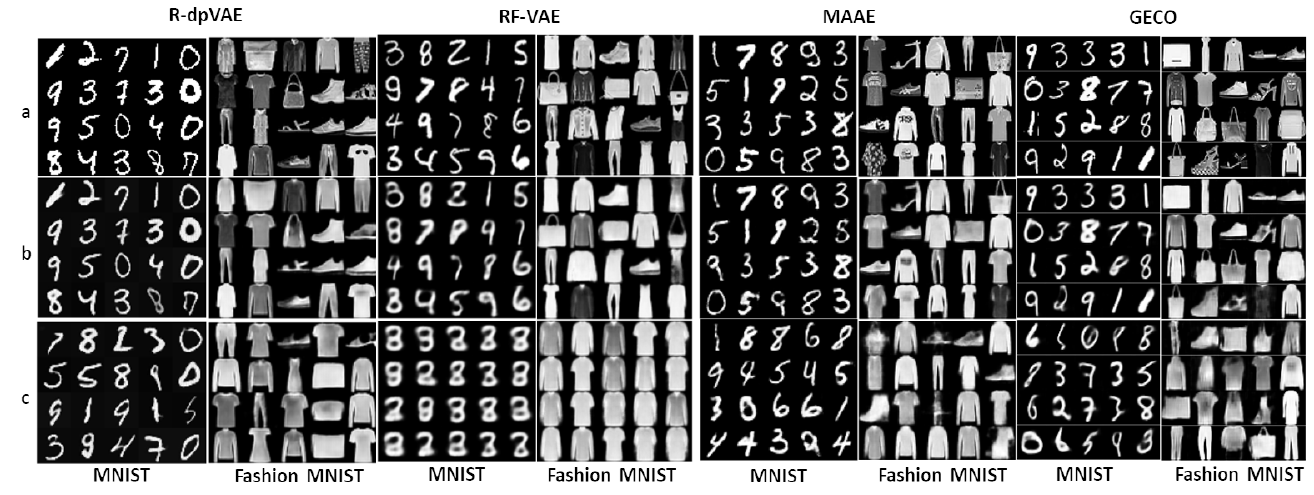}
    \caption{Results of the best performing models across \(L\) from Table~\ref{tab_fid_gen} for each method. We show (a) original images, (b) reconstructed images, and (c) randomly generated images (no cherry-picking).}
    \label{fig:best_model_image_results}
\end{figure}

Table~\ref{tab_rec_mse} summarizes the MSE of the testing (i.e., held-out) images for each model for all three datasets. While R-dpVAE does not achieve low MSE as other models, the overall performance in conjunction with the FID score indicates that R-dpVAE models generalized well with the use of calibrated decoders and relevance encoders. The use of fixed weight on the reconstruction term in models such as RF-VAE, MAAE, and GECO is hypothesized to cause lower MSE and higher FID scores. Figure~\ref{fig:best_model_image_results} shows the reconstructed and generated images for the best performing models cross different \(L\) values for each method.\\

\begin{table}[ht]
    \centering\scriptsize
    \begin{tabular}{lllllllll}
    \toprule
    \multicolumn{3}{c}{MNIST} & \multicolumn{3}{c}{Fashion MNIST} & \multicolumn{3}{c}{dSprites} \\
    \cmidrule(r){1-3} 
    \cmidrule(r){4-6} 
    \cmidrule(r){7-9} 
    Model   &   FID  & $L^*$   &   Model   &   FID   & $L^*$ &   Model   &   FID  & $L^*$ \\
    R-dpVAE $(L=16)$    &    6.57          & 6 &   R-dpVAE $(L=32)$   &\redb{48.61} &\redb{10} & R-dpVAE $(L=10)$   &  67.93 &7 \\
    R-dpVAE $(L=32)$    &   \redb{4.43} & \redb{11}  &   R-dpVAE $(L=64)$   &   52.68 &  26  & R-dpVAE $(L=15)$   &  \blueb{52.15} &\blueb{9}\\
    RF-VAE $(L=16)$  &   184.38  &  9&   RF-VAE $(L=32)$ &  258.37 &27& RF-VAE $(L=10)$ & 144.89 &8\\
    RF-VAE $(L=32)$  &   208.27  & 21&   RF-VAE $(L=64)$ &  272.87 &52& RF-VAE $(L=15)$ & 167.24 &12\\
    GECO $(L=16)$   &    29.21  & 11&   GECO $(L=32)$  &   69.67 &8 & GECO $(L=10)$  &  97.86 &5\\  
    GECO $(L=32)$   &    30.29  &  9&   GECO $(L=64)$  &   67.21 &10& GECO $(L=15)$  &  94.61 &5\\
    MAAE $(L=16)$   &    13.81  & 11 &   MAAE $(L=32)$ &  104.15 &6& MAAE $(L=10)$  & 105.89 &7\\
    MAAE $(L=32)$   &    13.04  & 11 &   MAAE $(L=64)$ &   79.18 &5& MAAE $(L=15)$  &  90.85 &11\\
    dpVAE $(L=16)$  &\blueb{4.62}  &\blueb{16} &   dpVAE $(L=32)$ &\blueb{52.08} & 32& dpVAE $(L=10)$ & \greenb{61.99} & \greenb{10} \\
    dpVAE $(L=32)$  &\greenb{5.53}  &\greenb{32} &   dpVAE $(L=64)$ &\greenb{52.38} & 64& dpVAE $(L=15)$ &  \redb{48.68} & \redb{15}\\
    VAE $(L=16)$    &     8.83  &16 &   VAE $(L=32)$   &   74.40 & 32& VAE $(L=10)$   &  82.667& 10 \\
    VAE $(L=32)$    &     9.65  &32 &   VAE $(L=64)$   &   70.18 & 64& VAE $(L=15)$   &  81.86 & 15\\
    \bottomrule \\
    \end{tabular}
     \caption{Generative metric FID (lower is better) for MNIST, Fashion MNIST, and dSprites and the identified bottleneck dimensionality by each model. FID = Frchet Inception Distance, \(L^*\) = Identified latent dimensionality. \textbf{\textcolor{red}{Best model}}, \textbf{\textcolor{blue}{second best model}}, \textbf{\textcolor[rgb]{0.0, 0.5, 0.0}{third best model}} }
    \label{tab_fid_gen}
\end{table}

\begin{table}[ht]
    \centering\scriptsize
    \begin{tabular}{llllll}
    \toprule
    \multicolumn{2}{c}{MNIST} & \multicolumn{2}{c}{Fashion MNIST} & \multicolumn{2}{c}{dSprites} \\
    \cmidrule(r){1-2} 
    \cmidrule(r){3-4} 
    \cmidrule(r){5-6} 
    Model   &   MSE(\(10e^{-2}\)) &   Model   &   MSE(\(10e^{-2}\))  &   Model   &   MSE(\(10e^{-2}\))  \\
    R-dpVAE $(L=16)$&   3.28 &   R-dpVAE  $(L=32)$  &1.21 & R-dpVAE 10 & 1.79 \\
    R-dpVAE $(L=32)$&   3.55           & R-dpVAE  $(L=64)$  &1.17 & R-dpVAE 15   & 1.35  \\
    RFVAE $(L=16)$  &\greenb{1.82} &   RFVAE  $(L=32)$ &\blueb{1.04}& RFVAE $(L=10)$ & 3.66 \\
    RFVAE $(L=32)$  &   1.83       &   RFVAE  $(L=64)$ &\redb{1.00}& RFVAE $(L=15)$ & 3.65 \\
    GECO $(L=16)$   &\redb{1.22} &   GECO  $(L=32)$  &\greenb{1.06}  & GECO $(L=10)$  & 2.45 \\  
    GECO $(L=32)$   &\blueb{1.24} &   GECO  $(L=64)$ &1.17 & GECO $(L=15)$  & 3.83 \\
    MAAE $(L=16)$   &   1.84  &   MAAE  $(L=32)$ &2.10 & MAAE $(L=10)$  &\redb{1.13} \\
    MAAE $(L=32)$   &   1.80   &   MAAE  $(L=64)$ &2.56 & MAAE $(L=15)$  &\greenb{1.28} \\
    dpVAE $(L=16)$  &   3.15  &   dpVAE  $(L=32)$ &1.24 & dpVAE $(L=10)$ &1.83 \\
    dpVAE $(L=32)$  &   3.22  &   dpVAE  $(L=64)$ &1.19 & dpVAE $(L=15)$ &9.67 \\
    VAE $(L=16)$    &   4.59   &   VAE  $(L=32)$   &1.36& VAE $(L=10)$   &2.95 \\
    VAE $(L=32)$    &   2.63   &   VAE  $(L=64)$   &1.35& VAE $(L=15)$   &\blueb{1.14} \\
    \bottomrule \\
    \end{tabular}
    \caption{Reconstruction error MSE of the testing sample for MNIST, Fashion MNIST, and dSprites and the identified bottleneck dimensionality by each model. MSE=Mean Squared Error, \(L^*\)=Identified latent dimensionality. \textbf{\textcolor{red}{Best model}}, \textbf{\textcolor{blue}{second best model}}, \textbf{\textcolor[rgb]{0.0, 0.5, 0.0}{third best model}}.}
    \label{tab_rec_mse}
\end{table}



\section{Conclusion}
Latent dimensionality mismatch can have a detrimental effect on the performance of deep generative models such as VAEs. We have introduced relevance encoding networks (RENs) to identify this mismatch and inform the model about the relevant bottleneck size. The RENs framework facilities training of VAEs using a unified probabilistic formulation to parameterize the data distribution and detect latent-intrinsic dimensionality mismatch. A key feature of the RENs framework is that it does not require any extra hyperparameter tuning for relevance determination and it provides a PCA-like ranking of the latent dimensions based on the learned, data-specific relevance. The proposed model is general and flexible to be incorporated in any state-of-the-art VAE-based models, including regularized variants of VAEs. Future directions include extending the formulation of RENs toward explainable VAEs.

\clearpage
\bibliography{refer}

\bibliographystyle{plain}

\appendix

\section{Appendix}

\subsection{Evidence Lower Bound for RENs}\label{elbo_proof}
From equation.\ref{ard_prior_1} and equation.\ref{ard_prior_2}, 
\begin{equation}\label{ard_prior_1_copy}
    \begin{split}
    p(\set{Z}|\boldsymbol{\alpha}) = \prod_{n=1}^N p(\sample{z}_n|\boldsymbol{\alpha});\\ p(\sample{z}_n|\boldsymbol{\alpha}) \sim \mathcal{N}(\sample{z}_n; \mathbf{0},\boldsymbol{\alpha}^{-1}\mathbb{I})
\end{split}
\end{equation}
\begin{equation}\label{ard_prior_2_copy}
    p(\boldsymbol{\alpha}) = \Gamma(a \mathbf{1}_L,b\mathbf{1}_L)
\end{equation}
Considering the graphical model in Figure.\ref{fig:ren_block_diagram}(d), the variational posterior is specified by \(q(\set{Z},\boldsymbol{\alpha}|\set{X})\)

\[q(\set{Z},\boldsymbol{\alpha}|\set{X}) = q_{\psi}(\boldsymbol{\alpha}|\set{Z},\set{X}) q_{\phi}(\set{Z}|\set{X}) \]
We begin with the defining the marginal likelihood of the training data,

\[p(\set{X})=\int_{\set{Z}} \int_{\boldsymbol{\alpha}} p(\set{X},\set{Z},\boldsymbol{\alpha}) \]
ELBO maximizes the marginal log likelihood of the training data: \(\log p(\set{X})\)
\[\log[p(\set{X})]=\log\left[\int_\set{Z} \int_{\boldsymbol{\alpha}} p(\set{X},\set{Z},\boldsymbol{\alpha}) \right]\]

\[=\log\left[\int_\set{Z} \int_{\boldsymbol{\alpha}} \frac{p(\set{X},\set{Z},\boldsymbol{\alpha}) q(\set{Z},\boldsymbol{\alpha}|\set{X})}{q(\set{Z},\boldsymbol{\alpha}|\set{X})} \right] \]

\[=\log\left[ \mathbb{E}_{q(\set{Z},\boldsymbol{\alpha}|\set{X})}\left[\frac{p(\set{X},\set{Z},\boldsymbol{\alpha})}{q(\set{Z},\boldsymbol{\alpha}|\set{X})}\right]\right]\]

Applying Jensens inequality

\[\log[p(\set{X})]\geq  \mathbb{E}_{q(\set{Z},\boldsymbol{\alpha}|\set{X})}\left[\log\left[\frac{p(\set{X},\set{Z},\boldsymbol{\alpha})}{q(\set{Z},\boldsymbol{\alpha}|\set{X})}\right]\right]\]

\[\geq  \mathbb{E}_{q(\set{Z},\boldsymbol{\alpha}|\set{X})}\left[ \log \left[ \frac{p_{\theta}(\set{X}|\set{Z},\boldsymbol{\alpha}) p(\set{Z}|\boldsymbol{\alpha})p(\set{\alpha})}{q_{\phi}(\set{Z}|\set{X})q_{\psi}(\boldsymbol{\alpha}|\set{Z},\set{X})} \right] \right]\]
 
\[\geq  \mathbb{E}_{q(\set{Z},\boldsymbol{\alpha}|\set{X})}\left[ \log\left[p_{\theta}(\set{X}|\set{Z},\boldsymbol{\alpha}) \right] \right] +\mathbb{E}_{q(\set{Z},\boldsymbol{\alpha}|\set{X})}\left[ \log \left[ \frac{p(\set{Z}|\boldsymbol{\alpha}) p(\boldsymbol{\alpha})}{q_{\phi}(\set{Z}|\set{X}) q_{\psi}(\boldsymbol{\alpha}|\set{Z},\set{X})} \right] \right]\]

The first term is the reconstruction loss:
\[p_{\theta}(\set{X}|\set{Z},\boldsymbol{\alpha}) = \prod_{n=1}^N p(\sample{x}_i|\sample{z}_i,\boldsymbol{\alpha})\]

\[\mathbb{E}_{q(\set{Z},\boldsymbol{\alpha}|\set{X})}\left[ \log\left[p_{\theta}(\set{X}|\set{Z},\boldsymbol{\alpha}) \right] \right] = \sum_{i=1}^N \mathbb{E}_{q(\sample{z}_i,\boldsymbol{\alpha}|\sample{x}_i)}\left[ \log\left[ p_{\theta}(\sample{x}_i|\sample{z}_i,\alpha) \right] \right]\]

The second term can be further simplified as follows,
\[\mathbb{E}_{q(\set{Z},\boldsymbol{\alpha}|\set{X})}\left[ \log \left[ \frac{p(\set{Z}|\boldsymbol{\alpha}) p(\boldsymbol{\alpha})}{q_{\phi}(\set{Z}|\set{X}) q_{\psi}(\boldsymbol{\alpha}|\set{Z},\set{X})} \right] \right]\]

\[= \mathbb{E}_{q(\set{Z},\boldsymbol{\alpha}|\set{X})}\left[ \log \left[ \frac{p(\set{Z}|\boldsymbol{\alpha})}{q_{\phi}(\set{Z}|\set{X})}\frac{p(\boldsymbol{\alpha})}{q_{\psi}(\boldsymbol{\alpha}|\set{Z},\set{X})} \right] \right]\]
\[ = \mathbb{E}_{q(\set{Z},\boldsymbol{\alpha}|\set{X})}\left[ \log \left[ \frac{p(\set{Z}|\boldsymbol{\alpha})}{q_{\phi}(\set{Z}|\set{X})}\right] + \log\left[ \frac{p(\boldsymbol{\alpha})}{q_{\psi}(\boldsymbol{\alpha}|\set{Z},\set{X})} \right] \right]\]

\[ = - \mathbb{E}_{q(\set{Z},\boldsymbol{\alpha}|\set{X})}\left[ \log \left[ \frac{q_{\phi}(\set{Z}|\set{X})}{p(\set{Z}|\boldsymbol{\alpha})}\right] + \log\left[ \frac{q_{\psi}(\boldsymbol{\alpha}|\set{Z},\set{X})}{p(\boldsymbol{\alpha})} \right] \right]\]

\begin{equation}
    \begin{split}
       \mathcal{L}(\phi, \theta, \psi) = \sum_{n=1}^N \mathbb{E}_{q(\sample{z}_i,\boldsymbol{\alpha}|\sample{x}_i)}\left[ \log\left[ p_{\theta}(\sample{x}_i|\sample{z}_i,\alpha) \right] \right] \\
       - \mathbb{E}_{q(\set{Z},\boldsymbol{\alpha}|\set{X})}\left[ \log \left[ \frac{q_{\phi}(\set{Z}|\set{X})}{p(\set{Z}|\boldsymbol{\alpha})}\right] + \log\left[ \frac{q_{\psi}(\boldsymbol{\alpha}|\set{Z},\set{X})}{p(\boldsymbol{\alpha})} \right] \right]
    \end{split}
\end{equation}

The ELBO for REN with VAEs: 
\begin{equation}
    \begin{split}
       \mathcal{L}(\phi, \theta, \psi) = \sum_{n=1}^N \mathbb{E}_{q(\sample{z}_i,\boldsymbol{\alpha}|\sample{x}_i)}\left[ \log\left[ p_{\theta}(\sample{x}_i|\sample{z}_i,\alpha) \right] \right] \\
         -\mathbb{E}_{q(\set{Z},\boldsymbol{\alpha} | \set{X})}[\log q_{\phi}(\set{Z}|\set{X}) ] 
        - \mathbb{E}_{q(\set{Z},\boldsymbol{\alpha} | \set{X})}[\log q_{\psi}(\boldsymbol{\alpha}|\set{X}.\set{Z})] \\ 
        +\mathbb{E}_{q(\set{Z},\boldsymbol{\alpha} | \set{X})}[\log p(\set{Z}|\boldsymbol{\alpha})] 
        + \mathbb{E}_{q(\set{Z},\boldsymbol{\alpha} | \set{X})}[\log p(\boldsymbol{\alpha})] 
    \end{split}
\end{equation}

For dpVAE \cite{bhalodia2020dpvaes}, please refer to the paper for the basic formulation. For dpVAE with REN the joint likelihood changes to, 
\[ p(\set{X},\set{Z},\boldsymbol{\alpha}) = p_{\theta}(\set{X}|\set{Z},\boldsymbol{\alpha})p(\set{Z}|\boldsymbol{\alpha})p(\boldsymbol{\alpha})\]
\[=p_{\theta}(\set{X}|\set{Z},\boldsymbol{\alpha})p(\set{Z}_{0}|\boldsymbol{\alpha}) \begin{vmatrix}\frac{\partial \set{Z}_0}{\partial \set{Z}}\end{vmatrix} p(\boldsymbol{\alpha})\]
\[=p_{\theta}(\set{X}|\set{Z},\boldsymbol{\alpha})p(g(\set{Z})|\boldsymbol{\alpha}) \begin{vmatrix}\frac{\partial g(\set{Z})}{\partial \set{Z}}\end{vmatrix} p(\boldsymbol{\alpha})\]

The second term for dpVAE with REN formulation becomes:

\[\mathbb{E}_{q(\set{Z},\boldsymbol{\alpha}|\set{X})}\left[ \log \left[ \frac{p(\set{Z}|\boldsymbol{\alpha}) p(\boldsymbol{\alpha})}{q_{\phi}(\set{Z}|\set{X}) q_{\psi}(\boldsymbol{\alpha}|\set{Z},\set{X})} \right] \right]\]

\[=\mathbb{E}_{q(\set{Z},\boldsymbol{\alpha}|\set{X})}\left[ \log \left[ \frac{p(g(\sample{z})|\boldsymbol{\alpha}) \begin{vmatrix}\frac{\partial g(\sample{z})}{\partial z}\end{vmatrix} p(\boldsymbol{\alpha})}{q_{\phi}(\set{Z}|\set{X}) q_{\psi}(\boldsymbol{\alpha}|\set{Z},\set{X})}\right] \right]\]
\[=- \mathbb{E}_{q(\set{Z},\boldsymbol{\alpha}|\set{X})} \left[ \log \left[\frac{q_{\phi}(\set{Z}|\set{X})}{p(g(\set{Z})|\boldsymbol{\alpha}) \begin{vmatrix}\frac{\partial g(\set{Z})}{\partial \set{Z}}\end{vmatrix}}   \right] + \log \left[\frac{q_{\psi}(\boldsymbol{\alpha}|\set{Z},\set{X})}{p(\boldsymbol{\alpha})}\right] \right]\]

\begin{equation}
    \begin{split}
        =- \mathbb{E}_{q(\set{Z},\boldsymbol{\alpha}|\set{X})} \left[ \log  q_{\phi}(\set{Z}|\set{X}) \right] - \mathbb{E}_{q(\set{Z},\boldsymbol{\alpha}|\set{X})} \left[ \log q_{\psi}(\boldsymbol{\alpha}|\set{X}.\set{Z}) \right]\\
        + \mathbb{E}_{q(\set{Z},\boldsymbol{\alpha} | \set{X})}[\log p(\boldsymbol{\alpha})] 
        + \mathbb{E}_{q(\set{Z},\boldsymbol{\alpha} | \set{X})} [\log p(g_{\eta}(\set{Z})) 
        + \log (\begin{vmatrix}\frac{\partial g_{\eta}(\set{Z})}{\partial \set{Z}}\end{vmatrix})]
    \end{split}
\end{equation}

The ELBO for REN with dpVAE: 

\begin{equation}
    \begin{split}
        \mathcal{L}(\phi, \theta, \psi) = \sum_{i=1}^N \mathbb{E}_{q(\sample{z}_i,\boldsymbol{\alpha}|\sample{x}_i)}\left[ \log\left[ p_{\theta}(\sample{x}_i|\sample{z}_i,\alpha) \right] \right] \\
         - \mathbb{E}_{q(\set{Z},\boldsymbol{\alpha}|\set{X})} \left[ \log  q_{\phi}(\set{Z}|\set{X}) \right] - \mathbb{E}_{q(\set{Z},\boldsymbol{\alpha}|\set{X})} \left[ \log q_{\psi}(\boldsymbol{\alpha}|\set{X}.\set{Z}) \right]\\
        + \mathbb{E}_{q(\set{Z},\boldsymbol{\alpha} | \set{X})}[\log p(\boldsymbol{\alpha})] 
        + \mathbb{E}_{q(\set{Z},\boldsymbol{\alpha} | \set{X})} [\log p(g_{\eta}(\set{Z})) 
        + \log (\begin{vmatrix}\frac{\partial g_{\eta}(\set{Z})}{\partial \set{Z}}\end{vmatrix})]
    \end{split}
\end{equation}
\subsection{ Training Algorithm}

\begin{algorithm}[H]
\caption{Training algorithm for RENs}\label{ren_algo}
\textbf{Input:} Training dataset \(\mathcal{X}\)\\
\textbf{Networks:} Encoder-\(\phi\), Decoder-\(\theta\), Relevance encoder-\(\psi\)\\
\textbf{Hyper-parameters:} VAE learning rate: \(lr_{VAE}\), REN learning rate: \(lr_{REN}\)\\
\textbf{Initialization:} \(\boldsymbol{\alpha} = \mathbf{1}^L\), \(\log \sigma = 0\), total number of epochs\(=\)epochs, number of burn-in epochs\(=\)burnin, relevance prior: \(p(\boldsymbol{\alpha}) = Gamma(1e^{-3},1e^{-4})\)\\
\For{e in range(epochs)}{   
    Shuffle \(\mathcal{X}\)\\
    \For{each batch in  \(\mathcal{X}\)}{
        Divide batch into \(r\) smaller batches: b\\
        \For{each b in batch}{
            Update \(\log\sigma,\phi\) and \(\theta\) using equation \ref{vae_eq} for VAE and equation \ref{dpvae_eq} for dpVAE with \(lr_{VAE}\)\\
        }
        \If{e\(>\)burnin}{
            
            Get new estimate of  \(\boldsymbol{\alpha} \)\\
            Update \(\log\sigma,\phi,\theta,\psi\) using equation \ref{vae_ren_eq} for VAE and equation \ref{dpvae_ren_eq} for dpVAE with \(lr_{REN}\)
        }
    }
}\end{algorithm}
\subsection{Toy Dataset Results}\label{appendix_toy_results}
\begin{figure}[ht!]
    \centering
    \includegraphics[scale=0.88]{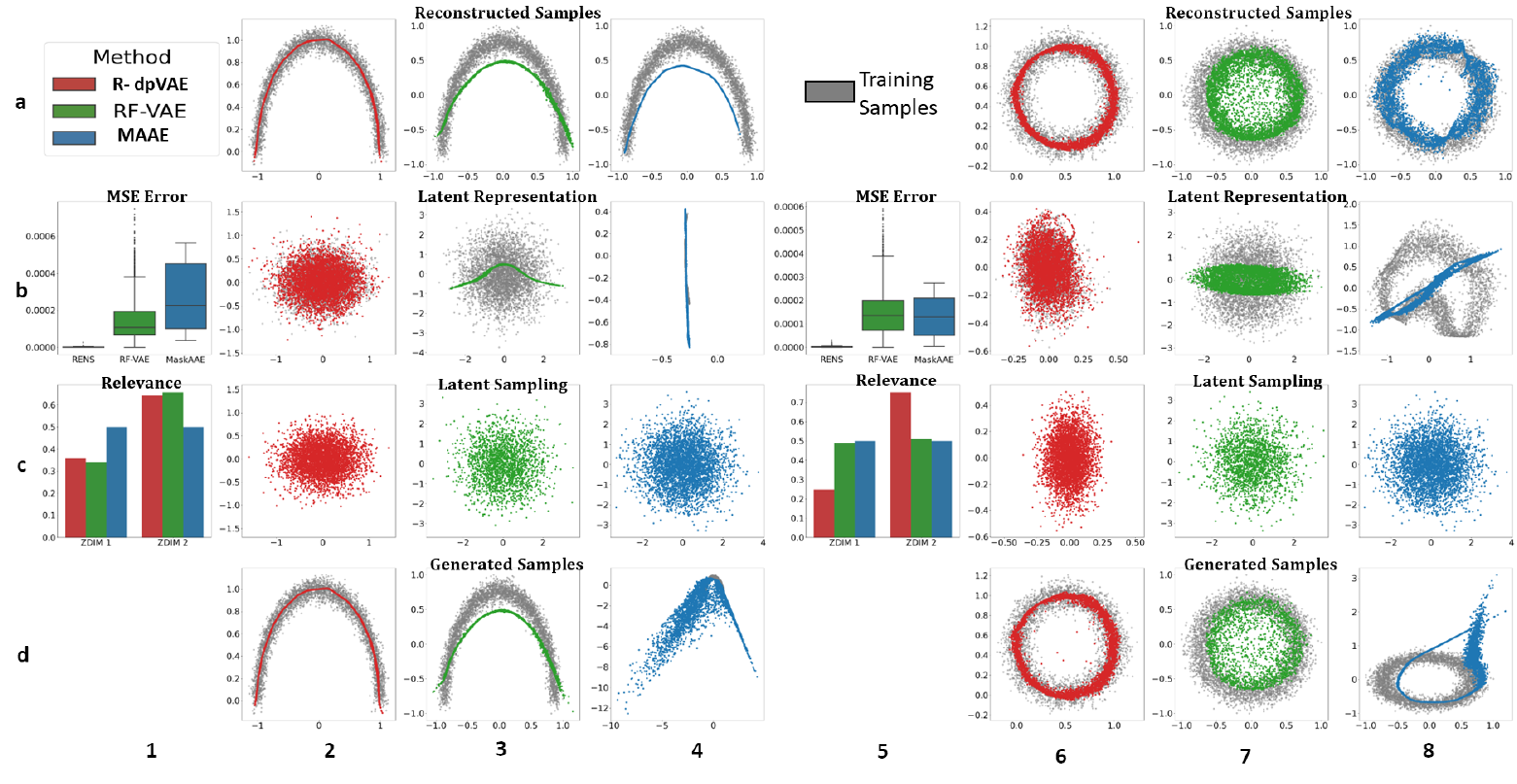}
    \caption{Sample reconstruction and sample generation outputs of R-dpVAE, RF-VAE, MaskAAE for one-moon and circle at 5\% and 7\% noise level.\\}
    \label{fig:one_moon_test}
\end{figure}

\begin{figure}[ht!]
    \centering
    \includegraphics[scale=0.88]{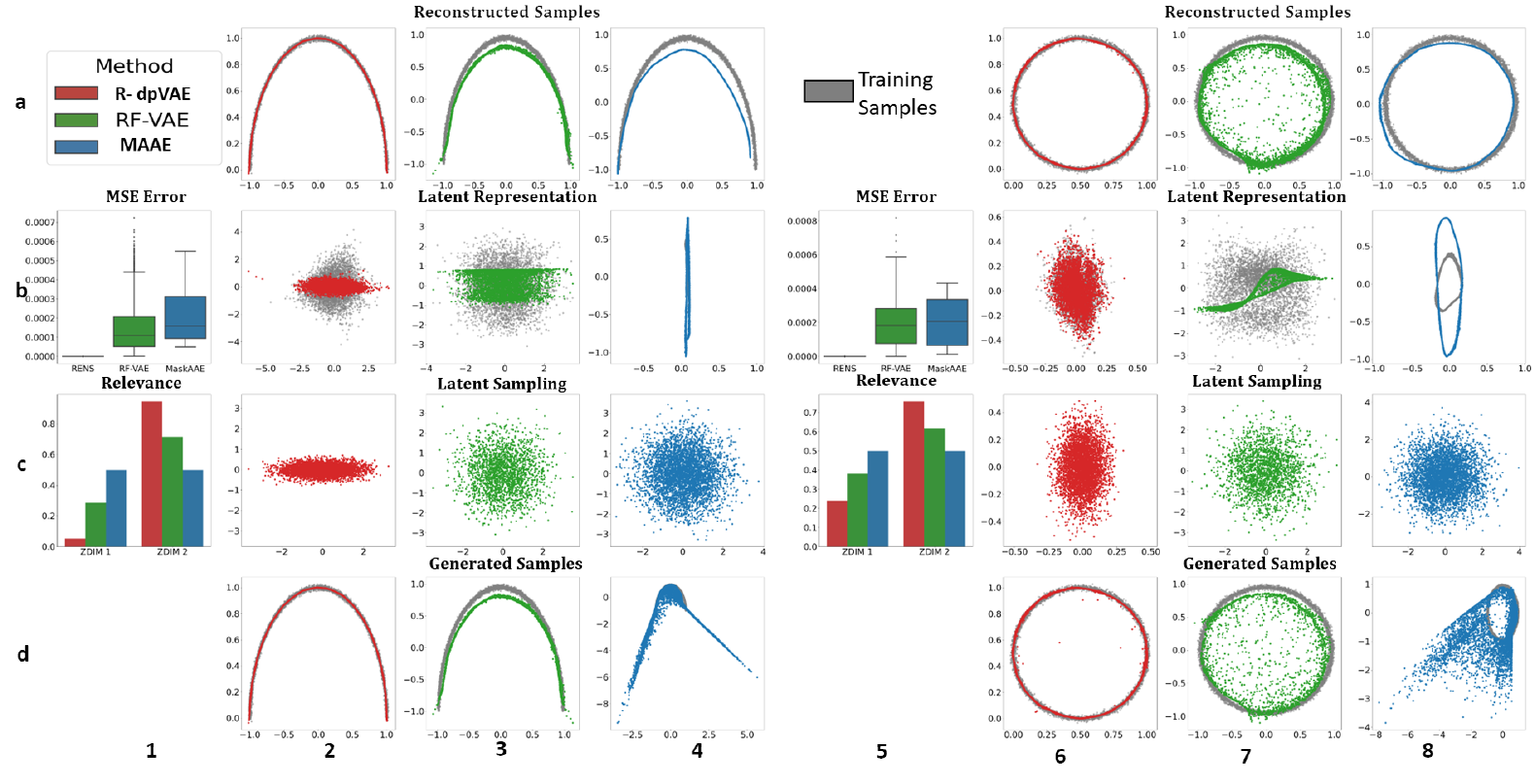}
    \caption{Sample reconstruction and sample generation outputs of R-dpVAE, RF-VAE, MaskAAE for one-moon and circle  at 1\% noise level.}
    \label{fig:circle_test}
\end{figure}

The results in Figure~\ref{fig:one_moon_test} and Figure~\ref{fig:circle_test} shows the results for higher noise levels. The observations are the same as section~\ref{image_experiments}. R-dpVAE is able to detect while maintaining low MSE even in the presence of noisy data.\\

\section{Broader Impacts}\label{impacts}
The paper proposes relevance encoding networks (RENs): a novel probabilistic VAE-based framework that uses the automatic relevance determination (ARD) prior in the latent space to learn the data-specific bottleneck dimensionality. The proposed method is an algorithmic improvement and does not have direct societal impacts. However, unsupervised representation learning using deep generative models needs to be evaluated critically. 

Unsupervised representation learning algorithms have emerged as alternatives for extracting meaningful and discriminative representations without labels. As the unsupervised representation learning algorithms rely solely on the training data, the model produced using such algorithms can only be as good as the quality of the training data used. The biases present in the training data can be propagated to the model. Hence, it is crucial to rigorously assess the quality of the training data, including biases, before using it for unsupervised algorithms. 

With the requirement of labeled data removed, the unsupervised representation learning algorithms have found applications in various domains like- computer vision, medical image analysis, natural language processing, etc. When using the algorithms for sensitive applications, it is paramount to introduce humans into the loop for decision-making to avoid automation bias.

\section{Architecture and Implementation Details}\label{implementation}

\begin{figure}[ht!]
    \centering
    \includegraphics[scale=0.8]{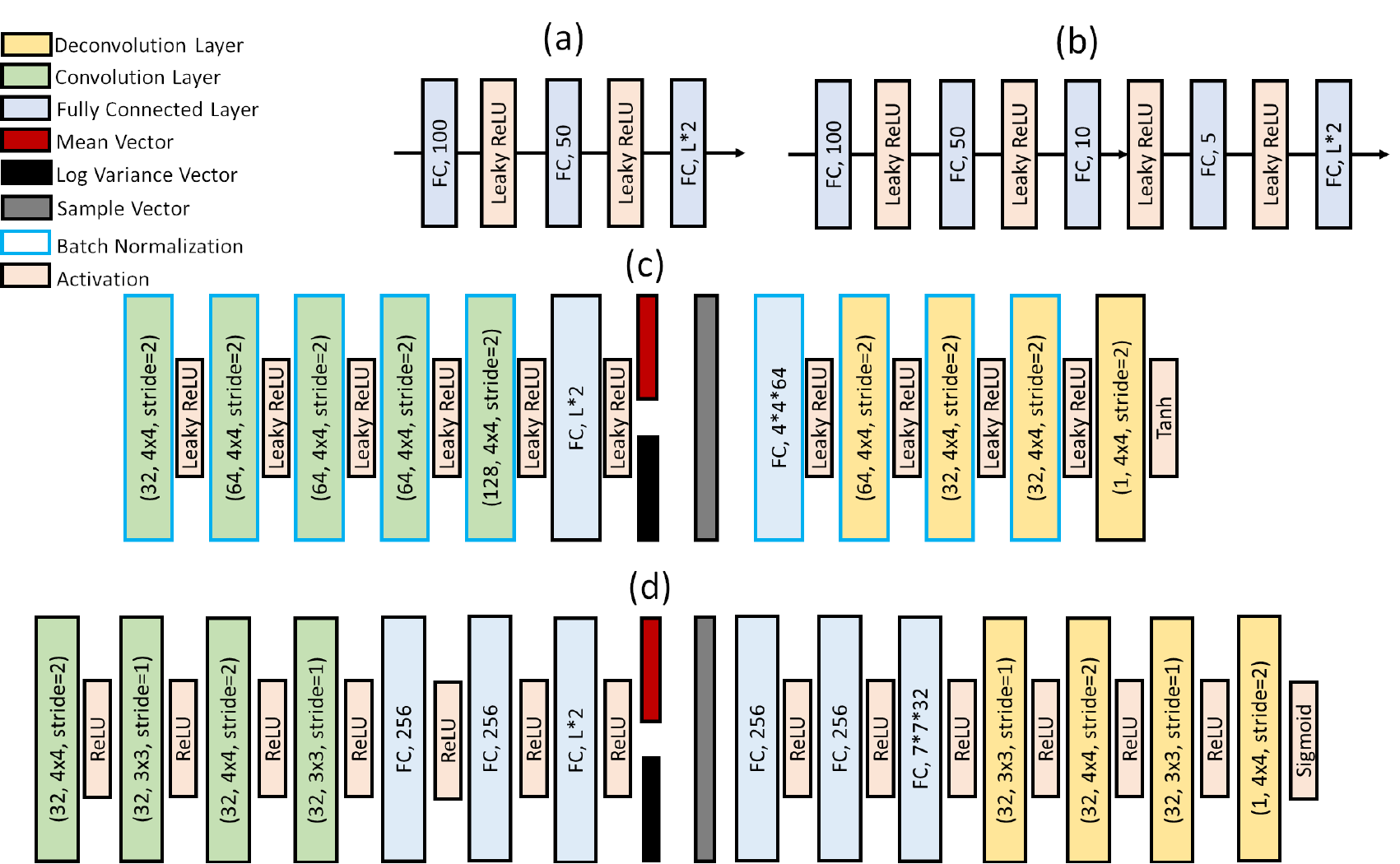}
    \caption{Encoder architecture for (a) one-moon, (b) circles dataset. Encoder and decoder architectures for (c) MNIST and Fashion MNIST, (d) dSprites }
    \label{fig:ren_decoder}
\end{figure}

\begin{figure}[ht!]
    \centering
    \includegraphics[scale=0.85]{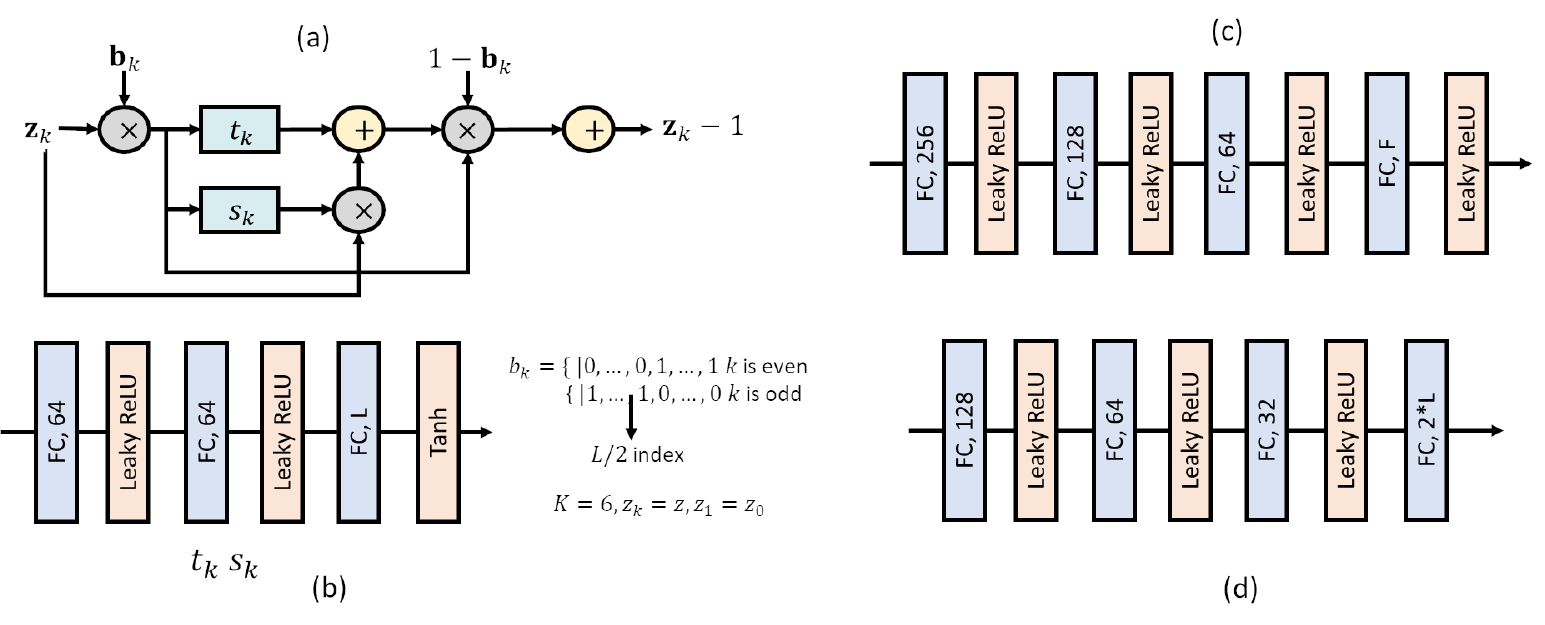}
    \caption{Architecture details of the decoupled prior (a) Individual affine coupling layer, (b) Individual scaling and transformation blocks. Architecture of the (c) feature extractor and (d) deep set aggregator of the relevance encoder. }
    \label{fig:coupling_layers}
\end{figure}

\subsection{RENs: Relevance Encoding Networks}
Figure~\ref{fig:ren_decoder} shows the architecture of the encoders and decoders used for all toy and image experiments. For toy datasets, the decoders were mirrored versions of the encoders. All the experiments used the same coupling layers and relevance encoder architecture shown in Figure~\ref{fig:coupling_layers}. The learning rate for VAEs using a smaller batch was set \(1e^{-3}\) and for an end to end training with the relevance encoders and a larger batch was set to \(1e^{-5}\). The large batch size for MNIST and Fashion MNIST was 100, and for dSprites, one-moon and circles were 128. In all the experiments, the larger batch was divided into four smaller batches (See Algorithm 1 for details). The number of samples in the training dataset of MNIST and Fashion MNIST was 60000, dsprites 663552, and toy experiments 4096. All image experiments were run for 100 epochs, and toy experiments were run for 1500 epochs. Seed used - 42 for all experiments. All experiments use the same gamma prior \(p(\boldsymbol{\alpha}) = \Gamma(1e^{-3}, 1e^{-4})\). The experiments were run on single GPU(NVIDIA GeForce RTX 2080 Ti). 

\subsection{RF-VAE: Relevance Factor VAE}
We used the author's implementation available at \url{https://github.com/ThomasMrY/RF-VAE}. 

\subsection{MAAE: Masked Adversarial Autoencoders}
We used the author's implementation available at:\\ 
\url{https://github.com/arnabkmondal/MaskAAE}

\subsection{GECO: VAE regularized with L0 ARM and GECO}
We used the code shared by the authors for the implementation after obtaining the consent for the use. 

\clearpage
\section{Results}\label{results}
\subsection{Image Datasets}
\begin{figure}[ht]
    \centering
    \includegraphics[scale=0.9]{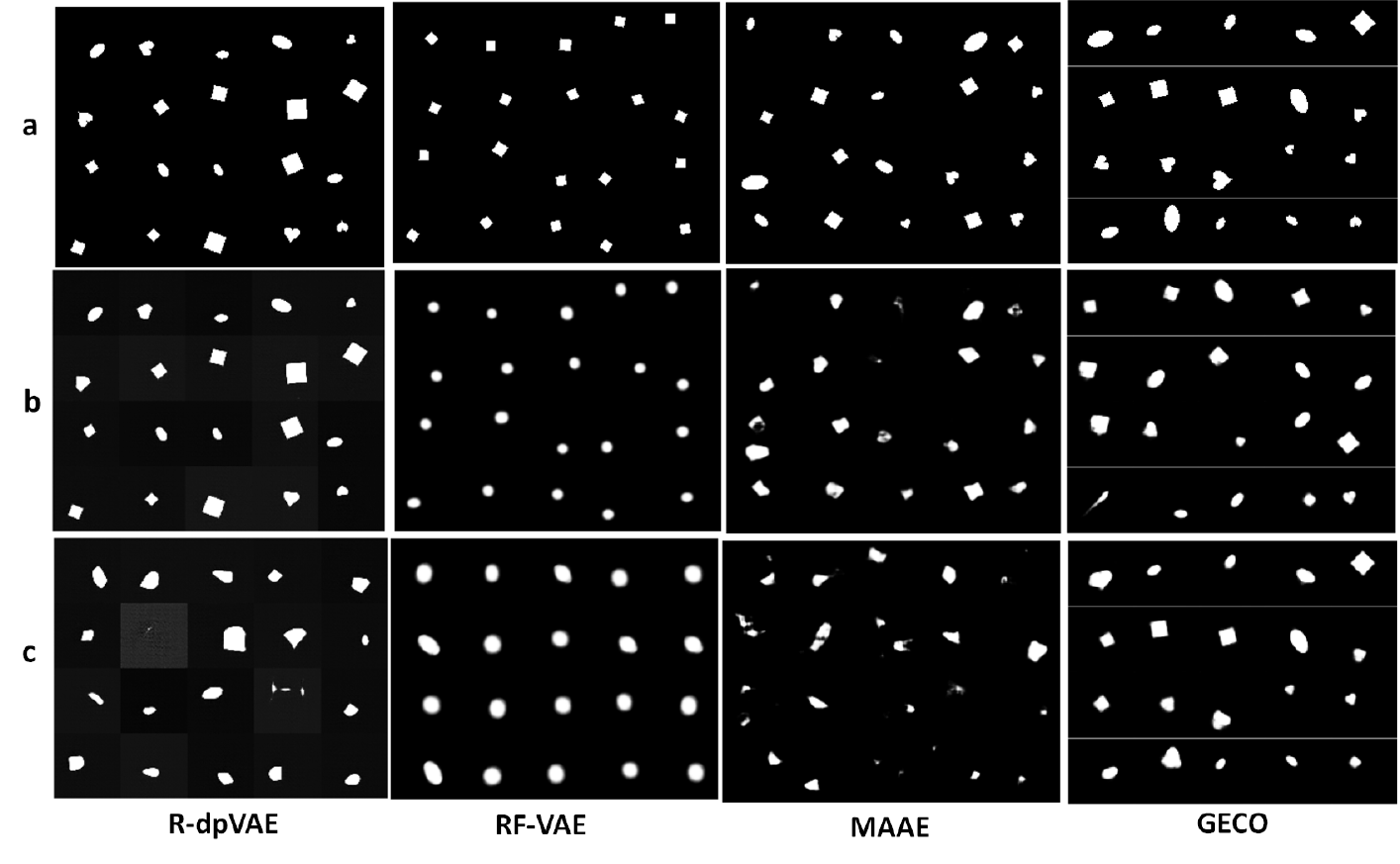}
    \caption{Results of the best performing models across \(L\) for dSprites from Table 1 for each method. We show (a) original images, (b) reconstructed images, and (c) randomly generated images (no cherry-picking).}
    \label{fig:dsprites}
\end{figure}

\begin{figure}[ht]
    \centering
    \includegraphics[scale=0.55]{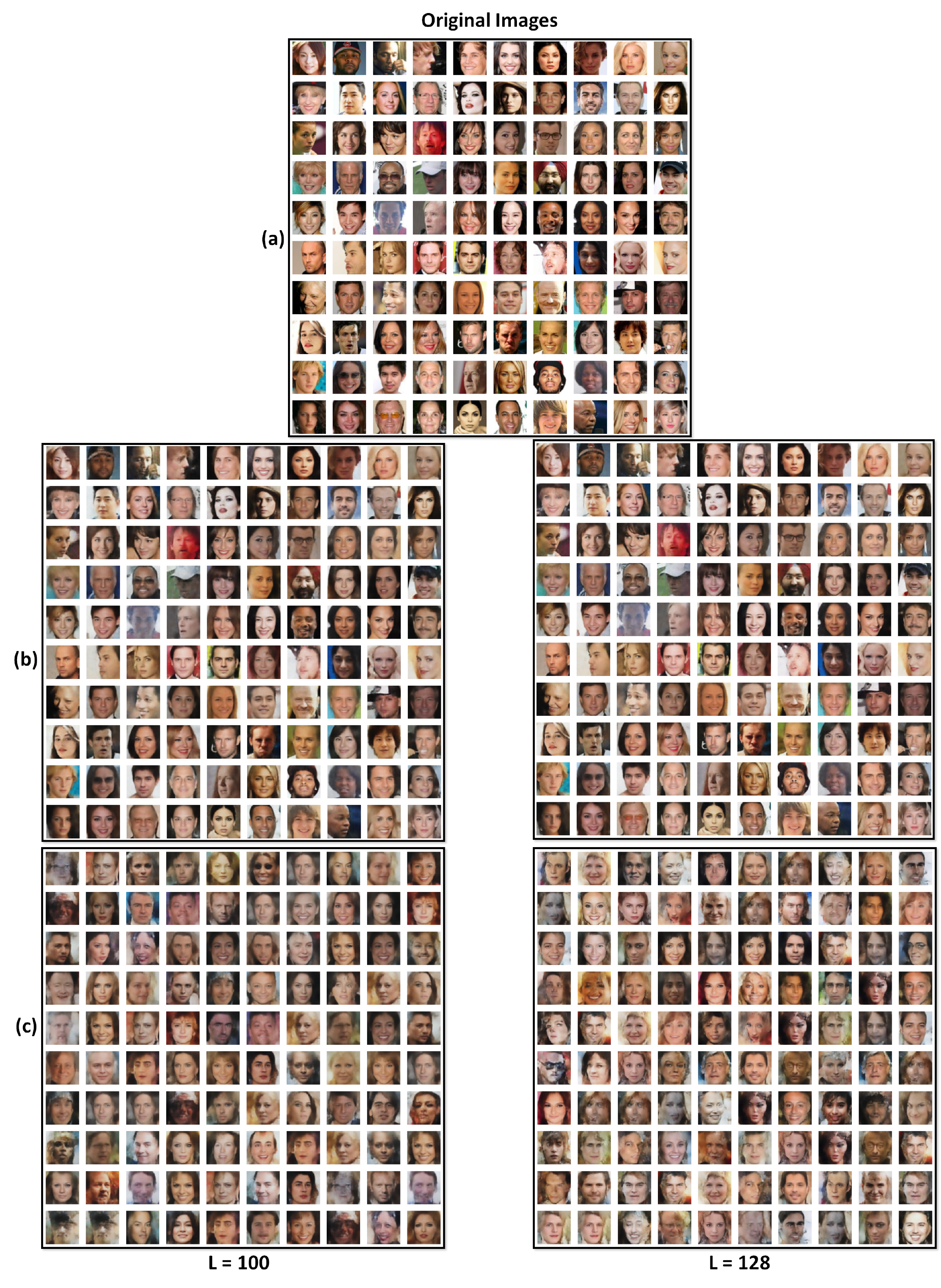}
    \caption{Results of R-dpVAE for CELEBA datasets. (a) Original images, (b) Reconstructed images, (c) Generated images (no cherry picking) for \textbf{(1) \(L = 100\) with estimated bottleneck size \(\mathbf{L^* = 37}\) and (2) \(L = 128\)  with estimated bottleneck size \(\mathbf{L^*=43}\)} }
    \label{fig:celeb_ren}
\end{figure}

\begin{figure}
    \centering
    \includegraphics{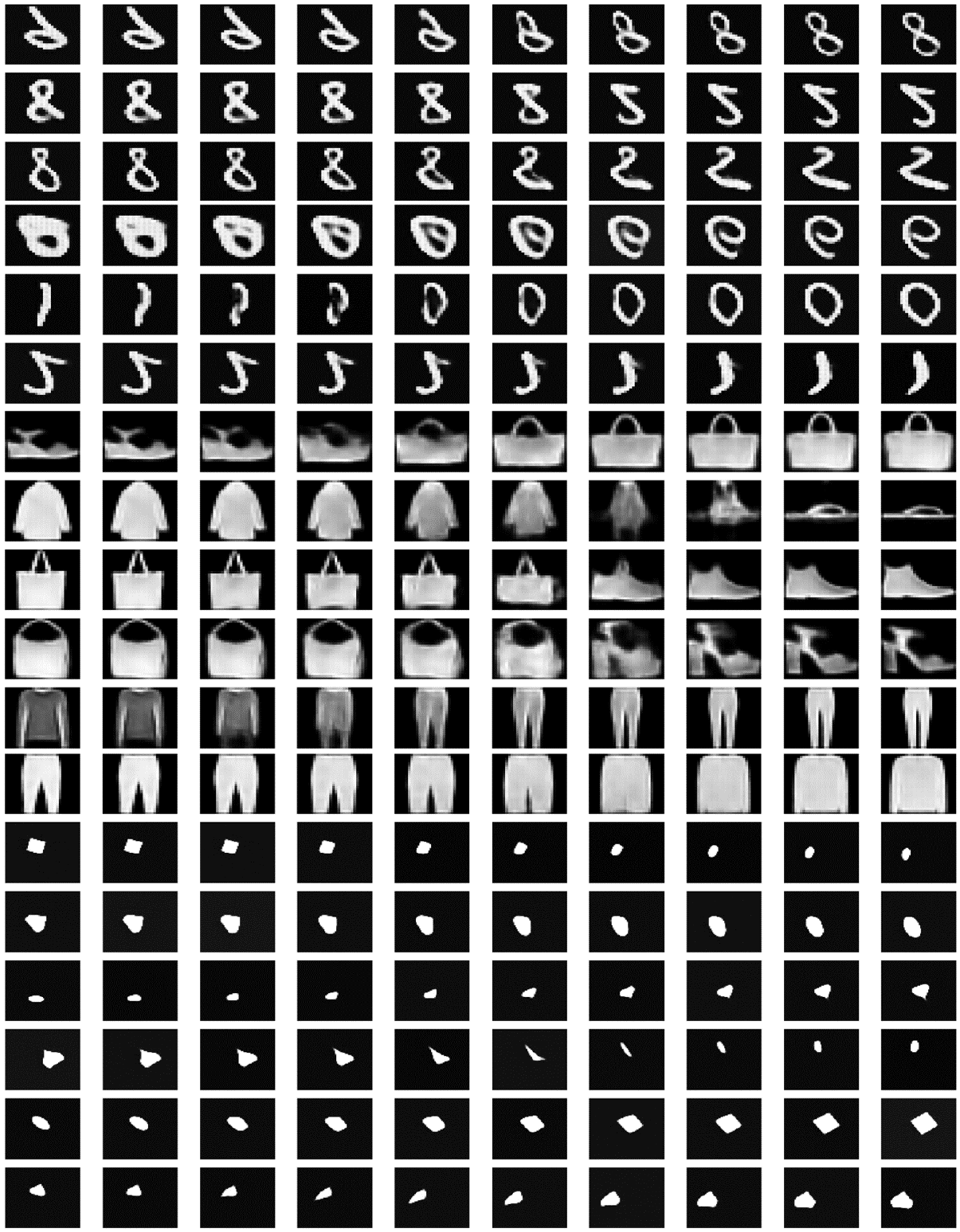}
    \caption{Each row represents a latent traversal from the image in the first column to the image in the last column of the best performing models of all  three datasets - R-dpVAE $(L=32)$ for MNIST and Fashion MNIST, R-dpVAE $(L=15)$ for dSprites. }
    \label{fig:mnist_traversal}
\end{figure}

\begin{figure}
    \centering
    \includegraphics[scale=0.6]{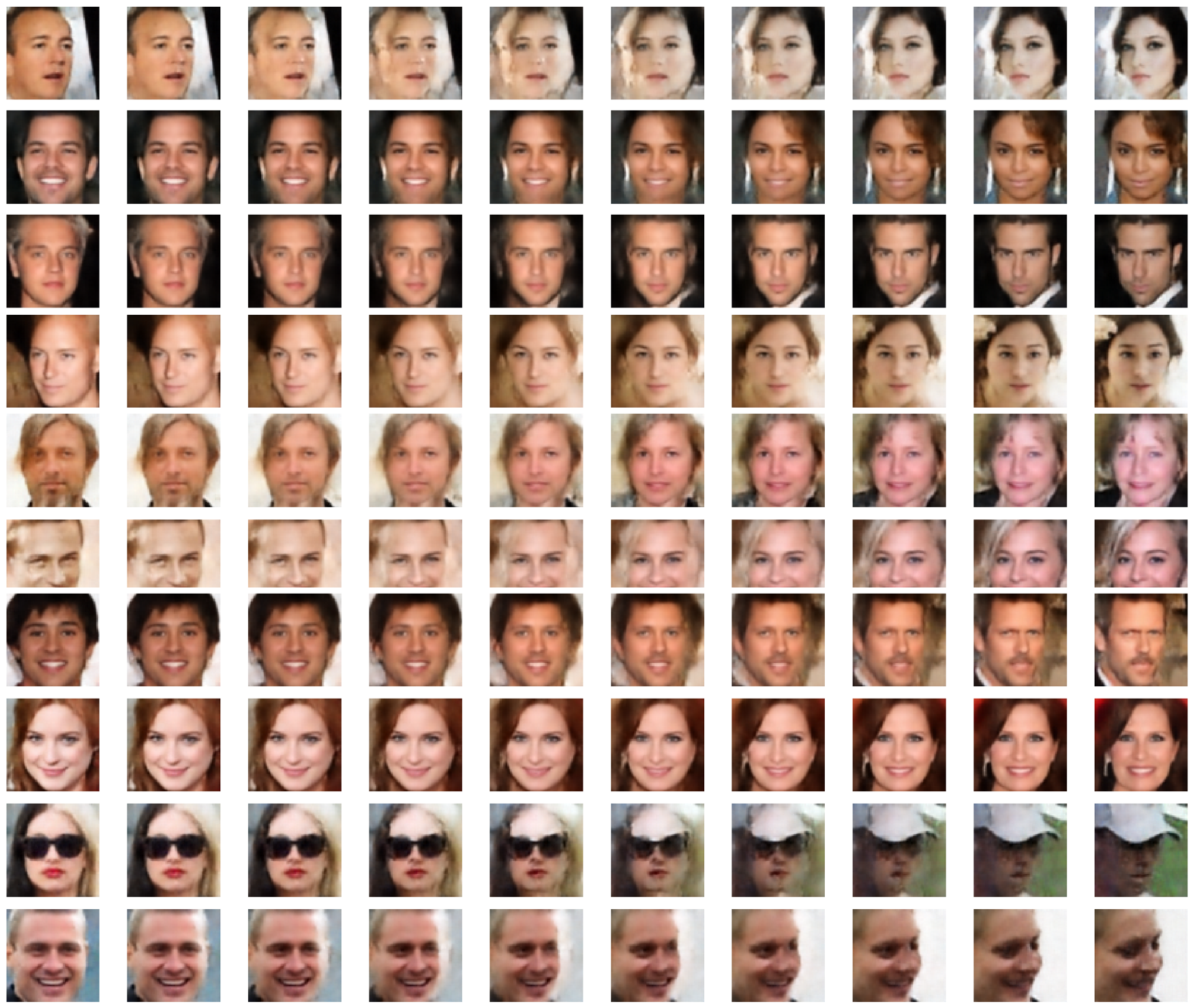}
    \caption{Each row represents a latent traversal from the image in the first column to the image in the last column for the CELEBA dataset using the R-dpVAE $(L=128)$ model.}
    \label{fig:celeba_traversal}
\end{figure}

In this section, we present more results for the image datasets. Figure~\ref{fig:dsprites} compares the performance of the various models on the dSprites dataset. Figure~\ref{fig:celeb_ren} shows the results of R-dpVAE on CELEBA dataset for \(L=100\) and \(L=128\). Figure~\ref{fig:mnist_traversal} and Figure~\ref{fig:celeba_traversal} shows the latent traversals of the R-dpVAE models. Each row represents a traversal from the image in the first column to the image in the second column. 

\clearpage

\clearpage
\section{Limitations}\label{limitations}
Relevance encoding networks have the same limitations as variational autoencoders (VAE). The images generated using RENs and VAE can be blurry and unrealistic compared to those generated using GANS. As RENs seek to maximize a lower bound on the log-likelihood of the generative models, RENs inherit the limitations of maximum-likelihood (ML) \cite{arjovsky2017towards}. Models like RENs and vanilla VAE optimize the (one-way) Kullback-Leibler (KL) divergence between the underlying data distribution and the model's distribution. Hence, do not penalize a model that generates data different from the training samples. Adversarial methods have been proposed in the literature that aims to alleviate this problem in VAE \cite{pu2017adversarial,hou2019improving}. Similar techniques could be utilized for RENs to improve the quality of generated images. 

\end{document}